%% file: main.tex
\documentclass[10pt,twocolumn,letterpaper]{article}

\usepackage{cvpr}              
\usepackage{algorithm}
\usepackage{algorithmic}
\usepackage{amsmath}
\usepackage{hhline}
 \usepackage{caption}
\usepackage{booktabs}
\usepackage{multirow}
\usepackage{tabularx}
\usepackage{placeins}
\usepackage{bbding}
\usepackage{pifont}
\usepackage{wasysym}
\usepackage{amssymb}
\usepackage{tikz}
\usepackage[table,svgnames]{xcolor}
\usepackage{pgfplots}
\usepackage{makecell}
\usetikzlibrary{positioning,calc,pgfplots.groupplots}
\pgfplotsset{compat=1.17}

\input{preamble}

\definecolor{cvprblue}{rgb}{0.21,0.49,0.74}
\usepackage[pagebackref,breaklinks,colorlinks,citecolor=cvprblue]{hyperref}

\def\paperID{137} 
\def\confName{3DV\xspace}
\def\confYear{2026\xspace}

\title{CamC2V: Context-aware Controllable Video Generation}

\author{
Luis Denninger\textsuperscript{1} \quad
Sina Mokhtarzadeh Azar\textsuperscript{1,2} \quad
Juergen Gall\textsuperscript{1,2} \\[2pt]
\textsuperscript{1}University of Bonn \quad
\textsuperscript{2}Lamarr Institute for Machine Learning and Artificial Intelligence \\
{\tt\small l\_denninger@uni-bonn.de \quad mokhtarzadeh@iai.uni-bonn.de \quad gall@iai.uni-bonn.de}
}

\begin{document}
\maketitle
\input{sec/0_abstract}    
\input{sec/1_intro}

\input{sec/2_related_works}

\input{sec/3_method}
\input{sec/4_experiments}

\input{sec/5_conclusion}

\section*{Acknowledgment}
This work has been supported the ERC Consolidator Grant FORHUE (101044724).
The authors gratefully acknowledge the granted access to the Marvin cluster hosted by the University of Bonn, as well as the Federal Ministry of Research, Technology and Space, the
Ministry of Culture and Science of the State of North
Rhine-Westphalia, the Ministry of Science, Research and
Arts of the State of Baden-Wurttemberg, the Bavarian
State Ministry of Science and the Arts and the Gauss
Centre for Supercomputing e.V. (GCS) for funding this
project by providing computing time on the Supercomputer
JUPITER at Julich Supercomputing Centre (JSC) of
Forschungszentrum Julich through the Gauss AI Compute
Competition.

{
    \small
    \bibliographystyle{ieeenat_fullname}
    \bibliography{main}
}
\input{sec/X_suppl}

\end{document}

%% file: preamble.tex
\usepackage[dvipsnames]{xcolor}


%% file: sec/0_abstract.tex
\begin{abstract}
Recently, image-to-video (I2V) diffusion models have demonstrated impressive scene understanding and generative quality, incorporating image conditions to guide generation. However, these models primarily animate static images without extending beyond their provided context. Introducing additional constraints, such as camera trajectories, can enhance diversity but often degrade visual quality, limiting their applicability for tasks requiring faithful scene representation. We propose CamC2V, a context-to-video (C2V) model that integrates multiple image conditions as context with 3D constraints alongside camera control to enrich both global semantics and fine-grained visual details. This enables more coherent and context-aware video generation. Moreover, we motivate the necessity of temporal awareness for an effective context representation. Our comprehensive study on the RealEstate10K dataset demonstrates a $24.09\%$ (FVD) improvement in visual quality and camera controllability. Our code is publicly available at: \href{https://github.com/LDenninger/CamC2V}{https://github.com/LDenninger/CamC2V}.
\end{abstract}

%% file: sec/1_intro.tex
\section{Introduction}\label{sec:intro}
Diffusion models have become a prominent approach for video generation producing high-quality videos based on user inputs.
To make such approaches attractive for digital content creation, controllability achieved through specific conditioning of the generations, 
like human poses~\cite{magicanimate, disco}, style~\cite{stylecrafter, stylemaster}, motion~\cite{motionctrl, videocomposer} or camera trajectories~\cite{cami2v, cameractrl, xu2024camcocameracontrollable3dconsistentimagetovideo} have been a widely studied topic.

While text-to-video (T2V) diffusion models like VideoCrafter~\cite{videocrafter} or CogVideoX~\cite{cogvideox} have full freedom over the visual design, more recent image-to-video (I2V) models employ an image to convey style and scene context.
Due to the typically short duration ($<2$ seconds) of the generated videos, the image provides sufficient context to define the scene to render.
With the ultimate objective of matching the generative quality and capabilities of traditional rendering engines,
these approaches still require further development to achieve a fine-grained control over style, motion and scene composition,
to allow for fully customizable video creation.

\begin{figure}
    \centering
    \includegraphics[width=0.9\linewidth]{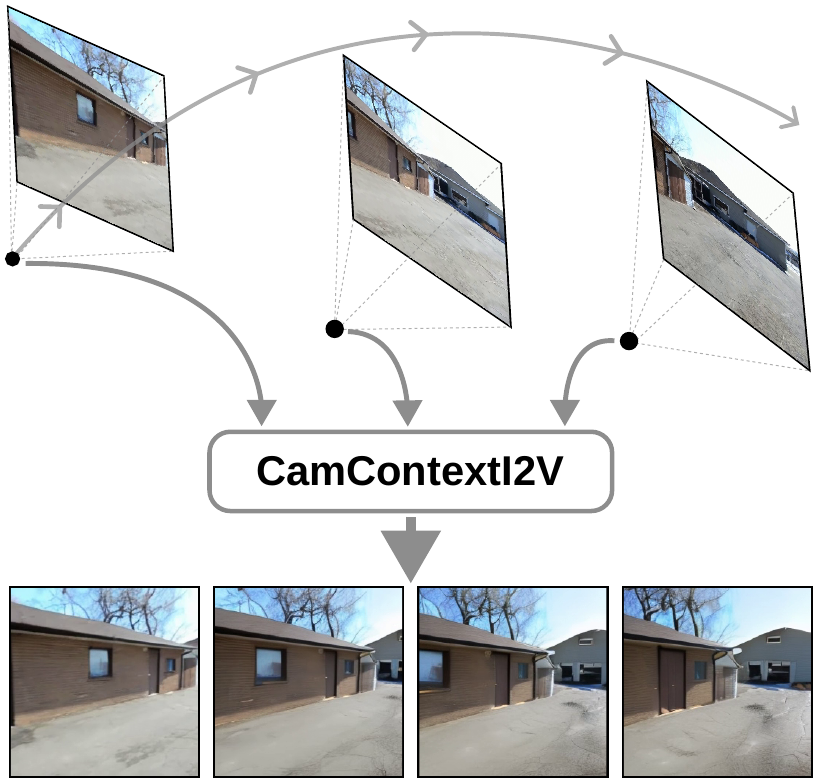}
    \caption{CamC2V performs context-aware generation provided a reference frame representing the initial frame $c_{img}$ and [1-4] additional views $c^{i}_{ctx}$ providing crucial context to the diffusion process missing in the reference frame.}
    \label{fig:title_fig}
\end{figure}

As illustrated in \cref{fig:title_fig}, the initial reference frame alone provides only limited context for the diffusion process.
Once the camera pans, the visual quality degrades and arbitrary interpretations of the scene by the diffusion model become evident.
To address this, we introduce CamC2V, a novel conditioning mechanism that allows users to supply multiple context views, ensuring a comprehensive definition of the scene in which the video is generated.

Our proposed \emph{Context-aware Encoder} integrates these context views into two complementary streams: a high-level semantic stream and a 3D-aware visual stream.
This dual-stream approach provides the diffusion model with both a global semantic context and a detailed pixel-level visual embedding.
By inserting 3D geometric constraints in the feature aggregation, we effectively retrieve important features from the context while filtering out irrelevant ones.
This allows our method to considerably enhance the visual coherence of existing approaches.
In summary, our key contributions are as follows.
\begin{itemize}
    \item We propose CamC2V, a camera-controllable context-aware diffusion model, which conditions the diffusion process on multiple context frames through a dual-stream encoder
    retrieving high-level semantic features and low-level visual cues from the context.
    \item We introduce a 3D-aware cross-attention mechanism leveraging epipolar constraints to effectively retrieve context from posed images.
    \item Our temporally-aware embedding strategy better aligns the context at different frame timesteps.
    \item Our method achieves a 24.09\% improvement in visual quality over the state-of-the-art methods on the RealEstate10K dataset. 
\end{itemize}

%% file: sec/2_related_works.tex
\section{Related Works}\label{sec:related_works}

\paragraph{Diffusion-based Video Generation.} 
Originally developed for image generation \citep{ddpm, latent_diff}, diffusion models have since demonstrated great success synthesizing high-quality videos~\cite{videodiffusion, stablevideodiffusion}.
Models such as SVD~\cite{stablevideodiffusion}, LAVIE~\cite{lavie} or VideoCrafter~\cite{videocrafter} have shown great success in distilling text-to-video (T2V) diffusion models from text-to-image (T2I) diffusion models by inserting temporal attention blocks modeling the added time dimension.
Building on top, models like DynamiCrafter~\cite{dynamicrafter}, Seine~\cite{seine} or I2vgen-XL~\cite{i2vgenxl} further fine-tune these models for image-to-video (I2V) generation showing impressive results. 



\paragraph{Camera-controllable Video Generation.} Concurrent work also focuses on adding camera control to diffusion models allowing the user to define the trajectory along a video is generated.
While initial work such as MotionCtrl~\cite{motionctrl}, AnimateDiff~\cite{animatediff} or Direct-a-Video~\cite{direct_a_video} model camera movements through camera-motion primitives, 
recent approaches such as CameraCtrl~\cite{cameractrl}, CamCo~\cite{xu2024camcocameracontrollable3dconsistentimagetovideo} or CamI2V~\cite{cami2v} directly insert the camera poses showcasing fine-grained camera control.
A key is the dense supervisory signal provided by pixel-wise camera rays represented as Plücker coordinates, which are encoded and inserted into the diffusion model in a ControlNet-like fashion~\cite{controlnet}.

CamCo and CamI2V further demonstrate that epipolar geometry can serve as an effective constraint in the information aggregation of vanilla attention mechanism.
While CamCo employs cross-attention to constraint the feature aggregation from the condition frame, CamI2V constrain the temporal self-attention itself to guide the diffusion process and thus improving the 3D consistency and camera trajectory.

\paragraph{Multi-Image Condition. } Large camera movements or longer generations result in multiple scenes being generated in one video which is insufficiently represented through a singular reference image
typically employed in concurrent image-to-video diffusion models~\cite{dynamicrafter,seine,i2vgenxl,cogvideox}.
Models like Gen-L-Video~\cite{gen_l_video}, MEVG~\cite{mevg} or VideoStudio~\cite{videostudio} explore the insertion of multiple text prompts to give a broader context across the temporal domain for longer video generation.
This is achieved by generating distinct short videos with different text conditions and optimizing the the noise between them either in a divide-and-conquer or auto-regressive setup to generate long consistent videos.

Recently approaches like ReCamMaster~\citep{bai2025recammaster} or TrajectoryCrafter~\citep{Yu_2025_ICCV} focus on conditioning the generative process on complete videos to recreate the video from another camera trajectory. While these methods effectively leverage the context of multiple images, their approaches heavily rely on the one-to-one mapping from condition and target frames. In contrast, our method only relies on loosely placed images that do not explicitely correspond to a timestamp in the target sequence.


%% file: sec/3_method.tex
\section{Preliminaries}\label{ssec:preliminaries}
Before we describe in Section \ref{ssec:problem_setting} our novel method, which enhances the context-awareness of pre-trained diffusion models by conditioning on multiple context views rather than a single reference frame, we briefly describe components of our baseline model, CamI2V, which extends DynamiCrafter~\cite{dynamicrafter}, a latent image-to-video diffusion model with camera pose conditioning.

\begin{figure*}[ht!]
    \centering
    \includegraphics[width=0.95\linewidth]{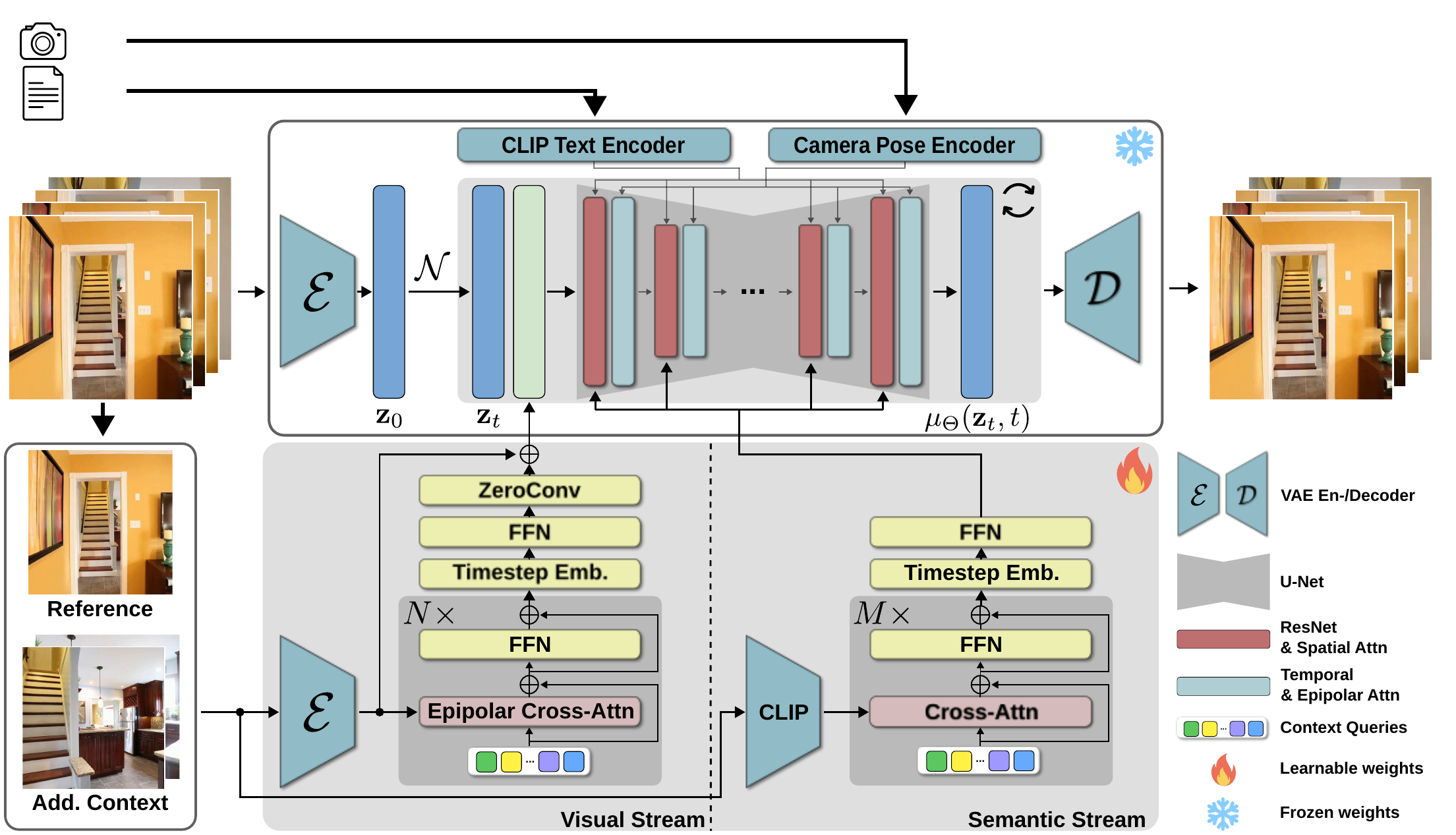}
    \caption{\textbf{CamC2V pipeline.} Our pipeline generates videos conditioned on a reference image, an optional text description and a camera trajectory encoded through a camera pose encoder conditioning. Additionally, the \emph{Context-aware Encoder} processes frames in two parallel streams, one providing pixel-level visual cues and the other a global context. The pixel-level stream employs epipolar attention to enforce 3D consistent feature aggregation. Finally, both stream are augmented with a timestep embedding to ensure timestep-wise conditioning of the diffusion process.}
    \label{fig:overview}
\end{figure*}

\paragraph{Latent Video Diffusion Models.} 
Latent video diffusion models learn a latent video data distribution by gradually reconstructing noisy latents $z_t$ sampled from a Gaussian distribution:
\begin{equation}
    q(z_t|z_{t-1}) = \mathcal{N}(z_t; \sqrt{1-\beta_t}, \beta_t\mathbf{I}),
\end{equation}
where hyperparameters $\beta_t$ determine the level of noise added at each timestep.
The latent space is defined through a pre-trained auto-encoder, e.g. a pre-trained VQGAN~\cite{vqgan} for DynamiCrafter, consisting of an encoder $\mathcal{E}$ and a decoder $\mathcal{D}$. Conditioned on a text condition $c_{\text{text}}$ and a reference image $c_{\text{img}}$, the diffusion model $\epsilon_{\Theta}$ is then trained to predict the noise $\epsilon$ at timestep $t\in{\mathcal{U}(0,T)}$ using a simple reconstruction loss:
\begin{equation}
    \min\limits_\theta \mathbb{E}_{t,\mathbf{x}\sim p_{data}, \epsilon\sim\mathcal{N}(\mathbf{0},\mathbf{I})} || \epsilon - \epsilon_\theta (\mathbf{x}_t, \mathbf{c}, t) ||_2^2.
\end{equation}
The diffusion model itself is typically implemented as a UNet, \eg a 3D-Unet~\cite{3dunet} in DynamiCrafter, where $\theta$ denotes the neural network's parameters.

\paragraph{Camera Conditioning.}
To incorporate camera control, CamI2V employs a dense supervisory signal using pixel-wise embeddings of camera rays, represented via Plücker coordinates.
Specifically, for each pixel $(u,v)$ the Plücker coordinates $P=(o\times d', d')$ are computed using the normalized ray direction $d' = \frac{d}{||d||}$ and the ray origin $o$ (the camera focal point).

The ray direction relative to a reference coordinate frame—such as the camera coordinate system of the initial frame—is derived from the intrinsics $\mathbf{K}$ and extrinsics $E=[\mathbf{R}|\mathbf{t}]$ as:
\begin{equation}
    d=\mathbf{R}\mathbf{K}^{-1} + \mathbf{t}.
\end{equation}
These embeddings are further encoded at multiple resolutions and integrated into the epipolar attention blocks inserted into the U-Net.


\section{Method}\label{ssec:problem_setting}


Image-to-video diffusion models generate videos based on a single reference frame $c_{img}$ and an optional text condition $c_{txt}$.
Additionally, camera-controlled diffusion models are conditioned on a camera trajectory $[P_{cam}^0,\dots,P_{cam}^{T}]$ allowing precise control of the camera 
view at each timestep. 
The reference frame does not always provide the necessary context corresponding to the camera trajectory. This can lead to insufficient visual quality of the generated frames.
In contrast, we propose a new scheme coined context-to-video which enhances the generation process with a rich context conveyed through additional context frames $c_{ctx}^0,\dots, c_{ctx}^{N}$ and their poses $P_{ctx}^0,\dots, P_{ctx}^{N}$



Our \emph{Context-aware Encoder}, shown in \cref{fig:overview}, extends DynamiCrafter's \emph{Dual-stream Image Injection} to support multiple image conditions.
Natively, it conditions the model at the pixel level by concatenating reference latents $z_{img}$ with noisy latents $z_t$ along the channel dimension, 
which restricts the generations to the narrow context provided by the reference image.
Additionally, to better guide the diffusion process, semantic features aggregated from CLIP-embedded image and text conditions are integrated layer-wise through spatial cross-attention.
To utilize the pre-trained generative capabilities of the diffusion model and refrain from fine-tuning large parts of the U-Net, we chose to inject our condition in those streams.

\paragraph{Semantic Stream.}
We adopt DynamiCrafter's query transformer $\mathcal{E}_{sem}$ to integrate cross-modal information from the CLIP-embedded reference image $\mathbf{F}_{img}$, the text condition $\mathbf{F}_{txt}$, and additional context frames $\mathbf{F}_{ctx}=[F_{ctx}^0,\dots,F_{ctx}^N]$. 
Specifically,  $\mathcal{E}_{sem}$ employs learnable latent query tokens $\mathbf{T}_{sem}$ to gather context across multiple layers of cross-attention and feed-forward networks, yielding a global representation:
\begin{equation}
    \mathbf{F}_{sem} = \mathcal{E}_{sem}([\mathbf{F}_{img}, \mathbf{F}_{txt}, \mathbf{F}_{ctx}], \mathbf{T}_{sem}).
\end{equation}
To preserve strong cross-modal context aggregation, we initialize  $\mathcal{E}_{sem}$ from DynamiCrafter’s \emph{ Dual-stream Image Injection} module and fine-tune it to handle multiple image conditions.

\paragraph{Visual Stream.}
While the semantic stream provides a well-suited global context representation, it lacks fine-grained visual details due to CLIP’s inherent training on visual-language alignment, which favors high-level representations of single entities.

To enhance context-aware generation, we integrate our visual condition directly into DynamiCrafter’s image conditioning.
Specifically, we embed the context frames $c_{ctx}^0,\dots, c_{ctx}^{N}$ into the latent space $\mathbf{Z}_{ctx} = [z_{ctx}^0,\dots, z_{ctx}^{N}]$ and introduce pixel-wise learnable context tokens $\mathbf{T}_{vis}\in\mathbb{R}^{T\times h \times w\times D}$.
The context tokens serve as queries in a query transformer, similar to the semantic stream, to aggregate timestep- and pixel-wise features from the latent context frames.

\paragraph{3D Awareness.} 
\begin{figure}[t]
    \centering
    \includegraphics[width=1.0\linewidth]{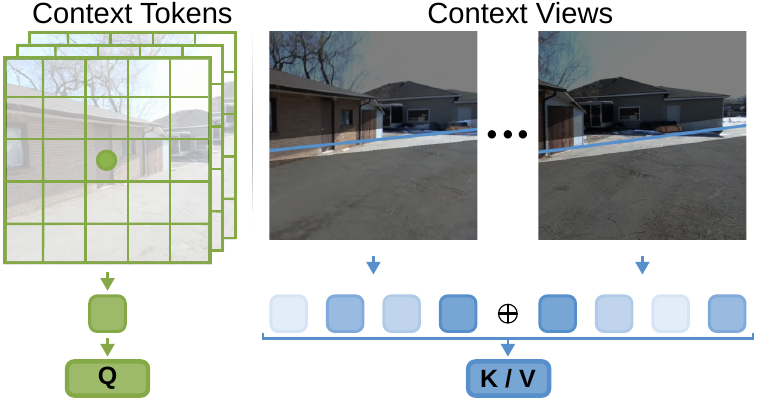}
    \caption{\textbf{Epipolar cross-attention.} Learnable context tokens act as queries to retrieve pixel-level features for each timestep from context views, masked according to epipolar lines to incorporate 3D geometric constraints.}
    \label{fig:epipolar_vis}
\end{figure}
To introduce 3D awareness, we employ an epipolar cross-attention mechanism which guides the feature aggregation to only consider potentially relevant features.
Specifically, each token $t_i\in \mathbf{T}_{vis}$, illustrated in \cref{fig:epipolar_vis}, describes a pixel $(u,v)$ at timestep $t$. 
Employing the provided camera pose $P_{cam}^t$ at the given timestep, we can compute the epipolar line $l_{ij} = Ax + Bx + C$ in each context view $c_{ctx}^j$.
Using the point-to-line distance:
\begin{equation}
    d(u',v')=\frac{[A,B,C]^\intercal \cdot [u',v',1]}{\sqrt{A^2+B^2}},
\end{equation}
we produce the epipolar mask $m\in\mathbb{R}^{Thw\times Nhw}$ masking out pixels $(u',v')$ with a distance larger than a threshold $\delta$, set to half of the diagonal of the latent feature space, in the cross-attention mechanism:
\begin{equation}
    \text{EpiCrossAttn(q,k,v,m)} = \text{softmax}\left(\frac{qk^\intercal}{\sqrt{d}}\odot m\right)v,
\end{equation}
where $q\in\mathbb{R}^{Thw\times D}$ describes the learnable context queries and $k,v\in\mathbb{R}^{Nhw\times D}$ the latent embedded context frames.

\paragraph{Temporal Awareness.}
The native pixel-level embedding of DynamiCrafter is agnostic to the timestep within the video as each timestep is provided with the same condition.
Thus, to further enforce the diffusion model to attend to context provided at specific timesteps, we found it advantageous to employ a sinusoidal timestep embedding.
In practice, we concatenate the timestep embedding to our context embeddings before forwarding it through a feed-forward network.

Finally, the visual stream of our \emph{Context-aware Encoder} maps a spatially distributed embedding represented through the latent embedding of posed views to a timestep-wise embedding:
\begin{equation}
    \mathbf{F}_{vis} = \mathcal{E}_{vis}(\mathbf{Z}_{ctx}, \mathbf{T}_{vis},  m).
\end{equation}
To retain the reference image as a strong anchor to the generation and smoothly insert the new condition, we employ a 3D zero-convolution which weighs
the usage of DynamiCrafter's native condition $z_{ref}$ and ours $\mathbf{F}_{vis}$ before adding them together.

\paragraph{Log-weighted Loss.}
Our \emph{Context-aware Encoder} injects crucial information, especially to later frames, that have limited context in the baseline methods leading to degrading visual quality.
To force the training process to focus on such frames that rely most heavily on our context‐aware conditioning, we apply a logarithmic re‐weighting to the standard reconstruction loss along the time axis. Specifically, for each frame $k$ in the sequence, we define:
\begin{equation}
    \mathcal{L} = \frac{\sum_{k=0}^{15}\log_{10}(k+1) \cdot || \varepsilon_k - \varepsilon_{\theta, k}(\mathbf{x_t}, \mathbf{c}, t)||_2^2}{\sum_{k=0}^{15}\log_{10}(k+1)}.
\end{equation}
This not only improves generative quality but also stabilizes the training, mitigating divergence in later stages of training.

%% file: sec/4_experiments.tex
\section{Experiments}\label{sec:experiments}
\begin{table*}[ht]
    \centering
    \small
    \begin{tabular}{l|ccc|ccc}
        \multirow{2}{*}{Method} 
        & \multicolumn{2}{c}{FVD~$\downarrow$} 
        & \multirow{2}{*}{MSE~$\downarrow$} 
        & \multirow{2}{*}{TransErr~$\downarrow$} 
        & \multirow{2}{*}{RotErr~$\downarrow$} 
        & \multirow{2}{*}{CamMC~$\downarrow$} \\
        & VideoGPT & StyleGAN & & & & \\
        \hline
        MotionCtrl    & 78.30 & 64.47 & 3654.54 & 2.89 & 2.04 & 4.34  \\
        \rowcolor{gray!15}CameraCtrl    & 71.22 & 58.05 & 3130.63 & 2.54 & 1.84 & 3.85  \\
        CamI2V        & \underline{71.01} & \underline{57.90} & \underline{2692.84} & \underline{1.79} & \underline{1.16} & \underline{2.58} \\
        \rowcolor{gray!15}Ours          & \textbf{53.90} & \textbf{45.36} & \textbf{2579.96} & \textbf{1.53} & \textbf{1.09} &  \textbf{2.29}  \\
    \end{tabular}
    
    
    \caption{\textbf{Quantitative comparison.} Compared against state-of-the-art camera-controlled diffusion models,
    our method achieves an improved video fidelity of $24.09\%$ in terms of FVD 
    The results were obtain using 25 DDIM steps with CFG set to 7.5, except for our method performing best with CFG set to 3.5}
    \label{tab:quant_comp}
\end{table*}
\begin{figure}[h]
    \centering
    \input{figures/mse_curve_new.tikz}
    \caption{\textbf{Frame-wise quantitative comparison.} 
    We compare the per-timestep MSE and SSIM against state-of-the-art methods.
    Due to the insufficient context provided by the reference frame, the visual quality degrades logarithmically as time progresses.}
    \label{fig:quant_comp_time}
\end{figure}
\begin{figure*}[t]
    \centering
    \input{figures/qual_figure2.tikz}
    \caption{\textbf{Qualitative comparison}. 
    Our method, provided with an additional context frame, overcomes the limited context of a single reference frame, improving visual quality beyond the reference frame. Zoom in for more details.}
    \label{fig:qual_comp}
\end{figure*}
\subsection{Setup}\label{sec:setup}
\paragraph{Dataset} The RealEstate10K~\cite{realestate10k} comprises approximately 70K video clips at 720p of static scenes depicting indoor and outdoor house tours. 
The clips are annotated with camera extrinsic and intrinsic values obtained through the ORB-SLAM2~\cite{orbslam2} pipeline. Additionally, we use the captions provided by the authors of CameraCtrl~\cite{cameractrl}.
The video clips are then center-cropped to a size of $256\times 256$ and clipped to short frames of length $16$ with a stride sampled between 1 and 10.

\paragraph{Metrics} We evaluate our method with respect to generative quality, the faithfullness to the provided context and the camera trajectory.
Firstly, to ensure improved video fidelity we report the Frechet Video Distance (FVD)~\cite{fvd} using the evaluation protocols from VideoGPT~\citep{yan2021videogptvideogenerationusing} and StyleGAN~\citep{karras2019stylebasedgeneratorarchitecturegenerative}.
To ensure the faithfulness with respect to the additional context, we evaluate the pixel-wise mean squared error (MSE) and the  Structural Similarity Index (SSIM)~\cite{ssim} independently for each timestep.

Finally, to examine the generated camera trajectory we follow the evaluation paradigm proposed by CameraCtrl and CamI2V.
Using GLOMAP~\cite{glomap}, we estimate the camera rotation $\tilde{R}_i$ and translation $\tilde{T}_i$ for each camera $i$ over $5$ trials and compute the independent rotation and translation errors, \emph{RotErr} and \emph{TransErr} respectively, as well as the combined element-wise error \emph{CamMC}:
\begin{align}
    \text{RotErr} &= \sum\limits_{i=1}^n \cos^{-1}\frac{\text{tr}(\tilde{R}_i R_i^T)-1}{2}, \\
    \text{TransErr} &= \sum\limits_{i=1}^n ||\tilde{T}_i-T_i||_2, \\ 
    \text{CamMC} &= \sum\limits_{i=1}^n ||[\tilde{R}_i|\tilde{T}_i] - R_i|T_i]||_2.
\end{align}
All metrics are computed on a subset consisting of videos extending over a duration of over $30$ seconds to ensure sufficient additional context to be sampled from 
and avoid sampling to close to the $16$ frame clip.

\paragraph{Implementation Details}
Initialized from CamI2V checkpoints, freezing all parameters except for our \emph{Context-aware Encoder},
we train for 50K iterations at a resolution of $256 \times 256$, using the Adam optimizer with a fixed learning rate of $1\times 10^{-4}$ and a batch size of $64$.
Using \textsc{Lightning} as our training framework with mixed-precision using DeepSpeed ZeRo-1 on 4 NVIDIA A100 GPUs, training takes approximately 7 days.
For comparison, we use the re-implementations of MotionCtrl~\cite{motionctrl} and CameraCtrl~\cite{cameractrl} provided by the authors of CamI2V.
We sample 1-4 context frames uniformly from the complete videos during training.

\subsection{Quantitative Comparison}\label{sec:quant_comp}
To show the effectiveness of the additional context provided by our method, we compare against several camera-controlled methods,
namely MotionCtrl~\cite{motionctrl}, CameraCtrl~\cite{cameractrl} and CamI2V~\cite{cami2v}.
\cref{tab:quant_comp} presents the comparison of our method against the baseline methods.
Our model achieves an improvement of $24.09\%$ in terms of the FVD score highlighting the effectiveness of added context for video generation.

To further evaluate the context-awareness of our method, we report the MSE in \cref{fig:quant_comp_time} between the generated videos and the ground-truth videos on a per-frame basis to assess
the improvement especially for later frames that typically lack sufficient context from the reference frame. Additionally, to assess the visual quality of each frame,
we compute the SSIM metric on each frame.

It is visible that the visual quality degrades logarithmically with the video length as the diffusion model lacks sufficient context.
Our method outperforms the baseline methods in both MSE and SSIM, especially for later frames.
This shows that providing the diffusion process with additional context can stabilize the generative quality over time.

Additionally, we investigate the accuracy of the generated camera trajectory with respect to the RotErr, TransErr and CamMC.
We observe a slightly improved rotational error  compared to CamI2V's, indicating an improved camera trajectory of our method.
As the evaluation pipeline, GLOMAP, used for estimating the camera trajectory matches keypoint features to simultaneously estimate the camera trajectory and reconstruct a 3D scene using bundle adjustment
and we do not train the camera encoder, nor the diffusion model itself, this improved camera trajectory is mainly linked to an improved 3D consistency and visual quality of the generated scene.
This demonstrates that the additionally provided context enforces are more faithful representation of the 3D scene.


\subsection{Qualitative Comparison}\label{sec:qual_comp}
\begin{figure*}[t!]
    \centering
    \input{figures/qual_result_sampling.tikz}
    \caption{\textbf{Qualitative results of different sampling strategies.} We generate samples conditioned on the furthest frame providing minimal context and the frame immediately following the video providing maximal context.
    Our method is able to reject unrelated features from the \textit{furthest} frame and only aggregate features from the \textit{end + 1} frame providing additional information to the diffusion process.}
    \label{fig:qual_comp_sampling}
\end{figure*}

\cref{fig:qual_comp} shows different samples from our method compared against CamI2V.
It is evident that the reference frame does not provide sufficient context for the generation past the first few frames.
This results in visually degrading image quality and unrealistic generations of the baseline method.

In contrast, our method is provided with an additional context frame sampled from a later timestep past the $16$ window frame that shows entities outside of the field of view of the reference frame or obstructed by obstacles.
Our method is able to comprehend the position of these entities in space and effectively embed it into the timestep-wise embedding resulting in these objects being placed at correct locations in later frames.
Moreover, it is visible, while the baseline method produces artifacts not visible in its condition, the extended context provides an additional constraint preventing unwanted artifacts.

%

\subsection{Ablation Studies}\label{sec:abl_studies}
\begin{table*}[h]
    \centering
    \small
    \begin{tabular}{cc|cc|cc|ccc}
         \multicolumn{2}{c|}{Multi-Cond.}
        & \multirow{2}{*}{Epipolar} 
        & \multirow{2}{*}{Time} 
        & \multicolumn{2}{c|}{FVD~$\downarrow$} 
        & \multicolumn{3}{c}{MSE~$\downarrow$} \\
         Pixel & Sem. &  &  & VideoGPT & StyleGAN 
        & Total & t=2 & t=16  \\
        \hline
          &  &  &  & 71.01  & 57.90  & 2792.84  & 758.38 & 4101.71 \\
          \hline
          \rowcolor{gray!15}\checkmark  &  & \checkmark  & \checkmark  & 76.00  & 63.40  & 2622.32 & \textbf{632.94} & 4141.67 \\
           & \checkmark & \checkmark  & \checkmark  & 70.44 & 59.56 & 2810.75  & 862.84 & 4225.31 \\
         \hline
          \rowcolor{gray!15}\checkmark & \checkmark &  &  & 63.817 & 54.13 & 2701.28 & 791.49 & 4127.12 \\
          \checkmark & \checkmark &  & \checkmark  & 61.61 & 52.04 & 2678.45 & 782.86 & 4102.77 \\
          \rowcolor{gray!15}\checkmark & \checkmark & \checkmark &  & \underline{58.15} & \underline{47.73}  & \underline{2642.69} & 753.36 &  $\mathbf{4014.67}$ \\
          \hline
          \checkmark & \checkmark & \checkmark  & \checkmark  & $\mathbf{53.90}$ & $\mathbf{45.36}$ & $\mathbf{2579.96}$ & \underline{668.60} & \underline{4076.78} \\
    \end{tabular}
    \caption{\textbf{Ablation studies.} We compare our design choices in different studies showing that our two-stream design complementarily embeds the context and guides the diffusion process. Adding epipolar attention and temporal embeddings to the \emph{Context-aware Encoder} equips it with explicit 3D and temporal awareness, improving context retrieval and further boosting performance.}
    \label{tab:ablation_studies}
\end{table*}
To thoroughly evaluate the impact of our design choices, we conducted several ablation studies.
The results are summarized in \cref{tab:ablation_studies}.

\paragraph{Semantic and visual stream.}

First, we examined the individual contributions of the semantic and visual streams to the diffusion process.
We trained two model variants, each utilizing only one stream to inject additional context.
Despite both variants being provided with an extended context, neither improved upon the baseline results.
This limited improvement likely stems from DynamiCrafter being originally trained under matching conditions.
In contrast, combining both semantic and visual streams significantly enhanced performance, highlighting their complementary interaction.

\paragraph{3D awareness.}
Next, we evaluated the effectiveness of our method's 3D awareness, achieved through the epipolar cross-attention mechanism.
Replacing epipolar cross-attention with standard (vanilla) cross-attention, allowing unrestricted feature aggregation from all tokens, still yielded a considerable improvement of $9.5$ FVD points over the baseline.
This model variant, still, demonstrates a significant improvement on the baseline by $9.5$ points in the FVD score
but fails to match the performance of the 3D-aware model variant.
This can be attributed to the model still leveraging the additional context for the generation but failing
to reject features from invalid positions, as seen in \cref{fig:qual_comp_sampling}, especially when context frames provide minimal additional information due to 
them being sampled from distant regions.

\paragraph{Temporal awareness.}
Further, we assess the effect of temporal embeddings integrated into semantic and visual streams.
Removing temporal embeddings results in a performance decline, although still outperforming CamI2V considerably.
The temporal embeddings, particularly within the visual stream, explicitly guide the temporal attention of the U-Net to properly interpret timestep-specific context.
Without this guidance, the epipolar cross-attention timestep-wise embedded context may be interpreted freely, resulting in impaired performance.
\begin{table}[h]
    \centering
    \small
    \renewcommand{\arraystretch}{1.3} 
    \begin{tabular}{lc|cc}
        Sample Range & FVD (VideoGPT) & MSE \\
        \hline
        $(\text{end}, -1]$ &  45.63 & 2579.96 \\
        $\text{end}+1$ & 44.21 & 2474.28\\
        Furthest & 48.52 & 2668.91 \\ 
    \end{tabular}
    
    \caption{\textbf{Condition sampling study.} To investigate the impact of different context views, we condition our method using different context sampling strategies. \textit{(end,-1]} represents the sampling strategy used through our evaluations, while \textit{end+1} provides context with the maximal amount of information and \textit{furthest} with the minimal amount of information.}
    \label{tab:sampling_studies}
\end{table}

\paragraph{Context sampling.}

Lastly, \cref{tab:sampling_studies} compares different sampling strategies for additional context views.
Our default method samples context frames from the interval $(\text{end}, -1]$ following the generated video.
Furthermore, we investigate two extremes: first, sampling a completely unrelated frame, the \emph{furthest} frame, as shown in \cref{fig:qual_comp_sampling}.
Our results show that this only slightly degrades the visual quality, indicating that our method effectively rejects unrelated features through the induced 3D awareness of the epipolar cross-attention.
Second, sampling a frame directly following the video, providing a maximal amount of information to the diffusion process.
This only slightly improves our method, showing that it can effectively gather context from loosely placed context views.
The qualitative results in \cref{fig:qual_comp_sampling} show that our \emph{context-aware encoder} effectively sorts out unrelated information
and provides the diffusion process only with the necessary context.

\subsection{Comparison to Novel View Synthesis Approach}\label{sec:nvs}

We finally compare our method against FrugalNeRF~\cite{frugalnerf}, a state-of-the-art novel view synthesis approach for sparse views. Such NeRF-based methods typically rely on an initial sparse 3D reconstruction step, limiting their applicability to diverse scenes. In a two-view setup, the COLMAP pipeline ~\cite{colmap}, typically employed in this step, only achieves a reliable registration in over $80\%$ of the cases if the frames distance is between $\sim 1 s$ and $\sim 8.5 s$, heavily limiting such approaches if the context frames are too close or too distant.

We train FrugalNeRF on $\sim100$ test scenes from the RealEstate10K dataset using the first and $17$th frame as training frames and the intermediate ones as test frames, similar to the \emph{End+1} evaluation setup of the context sampling study in \cref{sec:abl_studies}. Accordingly, we define the reference and context frame for our approach.

\cref{tab:nvs_eval} shows that our method slightly outperforms FrugalNeRF while being more broadly applicable, as it does not require a prior reconstruction step.
Moreover, because NeRF-based methods typically employ a test-time optimization training scheme, the time needed to learn and render the 3D representation ($1265.87s$) is substantially larger than the pure rendering time for our approach ($8.01s$), which can be directly applied to novel scenes after training.
See the supplementary material for additional details.
\begin{table}[h]
    \centering
    \small
    \setlength{\tabcolsep}{4pt} 
    \begin{tabular}{l|c c c c}
        \toprule
        Method & SSIM~$\uparrow$ & LPIPS~$\downarrow$ & PSNR~$\uparrow$ & Time (s) \\
        \midrule
             FrugalNeRF & 0.737 & 0.156 & 23.76  & 1265.87 \\
             Ours       & \textbf{0.741} & \textbf{0.128} & \textbf{24.32} & \textbf{8.01} \\
        \bottomrule
    \end{tabular}
    \caption{\textbf{Novel view synthesis comparison.} 
    We evaluate our model against a state-of-the-art sparse-view 3D reconstruction method that is geometrically grounded to faithfully represent the scene. 
    All methods are tested at a resolution of $512\times512$ using the first and 17th frames as conditioning signals.
    Our approach achieves slightly better image quality while requiring only a fraction of the compute time.
    }
    \label{tab:nvs_eval}
\end{table}

%% file: figures/mse_curve_new.tikz
\begin{tikzpicture}
  \begin{groupplot}[
      group style={
        group size=1 by 2,   
        vertical sep=1.2cm   
      },
      width=0.5\textwidth,
      height=3.5cm,
      axis line style={->,>=stealth},
      tick align=outside,
      tick style={draw=none,font=\scriptsize},       
      label style={font=\scriptsize},                
      title style={font=\small,yshift=-0.1cm},       
      xlabel style={font=\scriptsize,yshift=0.15cm}, 
      ylabel style={font=\scriptsize},
      grid=both,
      x grid style={dashed},
      grid style={line width=.1pt, draw=gray!10},
      major grid style={line width=.2pt, draw=gray!50},
      axis background/.style={fill=white},
      every axis plot/.append style={line join=round},
      clip=false,
    ]
    
    \nextgroupplot[
      title={MSE},
      xlabel={},
      ylabel={},
      xmin=1, xmax=16,
      xtick={2,4,6,8,10,12,14,16},
      ytick={0,1000,2000,3000,4000,5000,6000},
      scaled y ticks=base 10:-3,
      legend to name=combinedlegend,
      xticklabel style={font=\scriptsize,yshift=0.2cm}, 
      yticklabel style={font=\scriptsize,xshift=0.1cm}, 
      legend style={at={(0.5,-0.2)}, font=\footnotesize, draw=none, legend columns=4},
    ]
    \addplot[thick, color=green!60!black] coordinates {
      (1, 216.6927)  (2, 995.2471)   (3, 1666.7594) (4, 2249.2791)
      (5, 2754.0037) (6, 3190.1948)  (7, 3574.2678) (8, 3928.5886)
      (9, 4236.5605) (10,4506.5674)  (11,4726.8789) (12,4950.0103)
      (13,5143.5952)(14,5309.8335)  (15,5444.6147) (16,5579.5269)
    };
    \addlegendentry{MotionCtrl};
    
    \addplot[thick, color=magenta!70!black] coordinates {
      (1, 117.0798) (2, 701.5998)  (3, 1264.7761) (4, 1778.9578)
      (5, 2233.7036)(6, 2649.7874) (7, 3005.9875) (8, 3330.8518)
      (9, 3618.3342)(10,3880.7542) (11,4105.5210) (12,4326.2510)
      (13,4522.3579)(14,4692.7305) (15,4847.0962) (16,5014.2466)
    };
    \addlegendentry{CameraCtrl};

    \addplot[thick, color=blue!70!black] coordinates {
      (1,233.93) (2,686.74) (3,1162.45) (4,1593.42)
      (5,1989.53)(6,2328.85)(7,2623.39) (8,2885.31)
      (9,3121.79)(10,3332.89)(11,3513.56)(12,3678.88)
      (13,3840.41)(14,3986.46)(15,4122.33)(16,4258.20)
    };
    \addlegendentry{CamI2V};
    
    \addplot[thick, color=orange!80!black] coordinates {
      (1,186.4247)(2,563.4114)(3,930.0203)(4,1272.6462)
      (5,1572.5198)(6,1870.0884)(7,2127.1956)(8,2364.7185)
      (9,2546.5288)(10,2709.4810)(11,2863.4248)(12,3012.8516)
      (13,3141.4971)(14,3269.2310)(15,3365.2324)(16,3489.0010)
    };
    \addlegendentry{Ours};
    
    \nextgroupplot[
      title={SSIM},
      xlabel={Timestep \(t\)},
      ylabel={},
      xmin=1, xmax=16,
      xtick={2,4,6,8,10,12,14,16},
      xticklabel style={font=\scriptsize,yshift=0.2cm}, 
      yticklabel style={font=\scriptsize,xshift=0.1cm}, 
      ymin=0.2, ymax=1.0,
      ytick={0.2,0.4,0.6,0.8,1.0},
    ]
    \addplot[thick, color=blue!70!black] coordinates {
      (1,0.85499) (2,0.68128) (3,0.60746) (4,0.55839)
      (5,0.52076) (6,0.49178) (7,0.46908) (8,0.44948)
      (9,0.43365) (10,0.42012)(11,0.40831)(12,0.39806)
      (13,0.38980)(14,0.38173)(15,0.37601)(16,0.37118)
    };
    \addplot[thick, color=orange!80!black] coordinates {
      (1,0.85229) (2,0.68685) (3,0.61456) (4,0.56700)
      (5,0.53214) (6,0.50452) (7,0.48307) (8,0.46375)
      (9,0.44957) (10,0.43809)(11,0.42567)(12,0.41790)
      (13,0.40858)(14,0.40077)(15,0.39554)(16,0.39093)
    };
    \addplot[thick, color=green!60!black] coordinates {
      (1,0.84895) (2,0.60455) (3,0.52232) (4,0.47650)
      (5,0.44593) (6,0.42271) (7,0.40515) (8,0.39031)
      (9,0.37806) (10,0.36694)(11,0.35797)(12,0.35035)
      (13,0.34459)(14,0.33920)(15,0.33555)(16,0.33270)
    };
    \addplot[thick, color=magenta!70!black] coordinates {
      (1,0.85773) (2,0.66464) (3,0.58188) (4,0.52967)
      (5,0.49271) (6,0.46453) (7,0.44207) (8,0.42316)
      (9,0.40806) (10,0.39494)(11,0.38436)(12,0.37489)
      (13,0.36758)(14,0.36130)(15,0.35674)(16,0.35268)
    };
    
  \end{groupplot}
  \node at ($(current bounding box.south)!0.5!(current bounding box.south)$)
    [anchor=north, yshift=3.5pt]
    {\pgfplotslegendfromname{combinedlegend}};
\end{tikzpicture}

%% file: figures/qual_figure2.tikz
\begin{tikzpicture}[x=1.5cm, y=-1.5cm, scale=1.2]
    \newcommand{\qualImageSize}{1.7cm}
    \def\colgap{0.02} 
    \def\gapscale{0.9}
    \tikzset{myNodeStyle/.style={draw, inner sep=0pt, outer sep=0pt, minimum size=\qualImageSize}}

    \def\headerY{0.45}

    \node[anchor=south, inner sep=0pt, font=\footnotesize\bfseries] at (1,\headerY) {\textbf{Reference}};
    \node[anchor=south, inner sep=0pt, font=\footnotesize\bfseries] at (2,\headerY) {\textbf{Context}};
    
    \node[rotate=90, font=\footnotesize\bfseries] at (0.4,1) {\textbf{GT}};
    \node[rotate=90, font=\footnotesize\bfseries] at (0.4,2) {\textbf{CamI2V}};
    \node[rotate=90, font=\footnotesize\bfseries] at (0.4,3) {\textbf{Ours}};
    \node[rotate=90, font=\footnotesize\bfseries] at (0.4,4.2) {\textbf{GT}};
    \node[rotate=90, font=\footnotesize\bfseries] at (0.4,5.2) {\textbf{CamI2V}};
    \node[rotate=90, font=\footnotesize\bfseries] at (0.4,6.2) {\textbf{Ours}};
    
    \node[myNodeStyle] (img-1-1) at (1,1) {%
      \includegraphics[width=\qualImageSize]{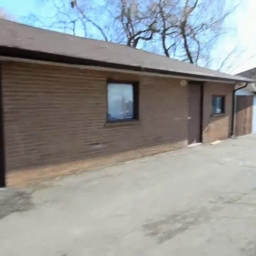}};
    \node[myNodeStyle] (img-1-3) at ({3+\colgap},1) {%
      \includegraphics[width=\qualImageSize]{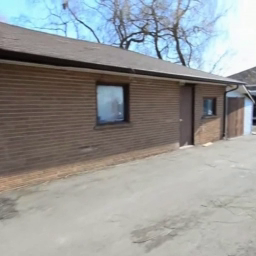}};
    \node[myNodeStyle] (img-1-4) at ({4+\colgap},1) {%
      \includegraphics[width=\qualImageSize]{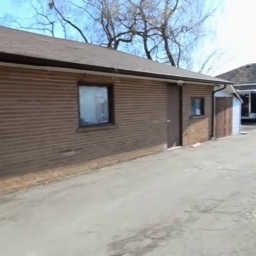}};
    \node[myNodeStyle] (img-1-5) at ({5+\colgap},1) {%
      \includegraphics[width=\qualImageSize]{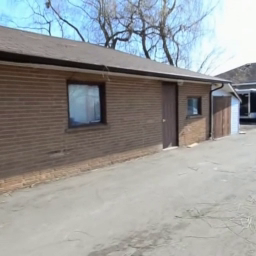}};
    \node[myNodeStyle] (img-1-6) at ({6+\colgap},1) {%
      \includegraphics[width=\qualImageSize]{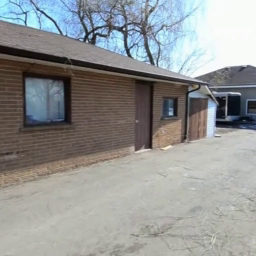}};
    \node[myNodeStyle] (img-1-7) at ({7+\colgap},1) {%
      \includegraphics[width=\qualImageSize]{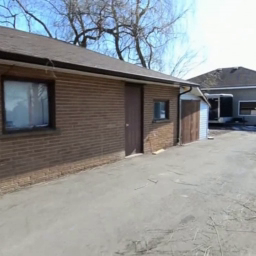}};
    \node[myNodeStyle] (img-1-8) at ({8+\colgap},1) {%
      \includegraphics[width=\qualImageSize]{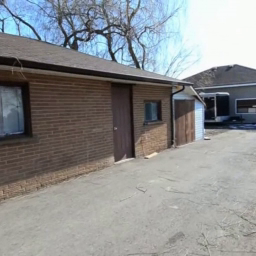}};
    
    \node[myNodeStyle] (img-2-1) at (1,2) {%
      \includegraphics[width=\qualImageSize]{images/qual_results/raw_data/ours/acdcd593a8f50eb1/ground_truth/output_0001.png}};
    \node[myNodeStyle] (img-2-3) at ({3+\colgap},2) {%
      \includegraphics[width=\qualImageSize]{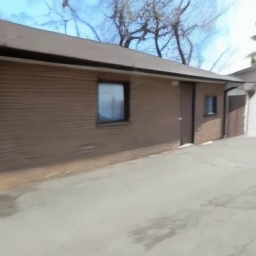}};
    \node[myNodeStyle] (img-2-4) at ({4+\colgap},2) {%
      \includegraphics[width=\qualImageSize]{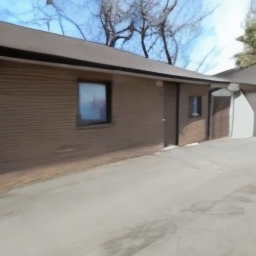}};
    \node[myNodeStyle] (img-2-5) at ({5+\colgap},2) {%
      \includegraphics[width=\qualImageSize]{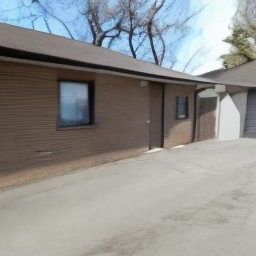}};
    \node[myNodeStyle] (img-2-6) at ({6+\colgap},2) {%
      \includegraphics[width=\qualImageSize]{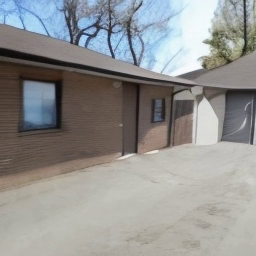}};
    \node[myNodeStyle] (img-2-7) at ({7+\colgap},2) {%
      \includegraphics[width=\qualImageSize]{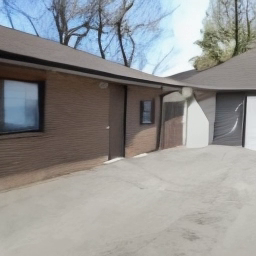}};
    \node[myNodeStyle] (img-2-8) at ({8+\colgap},2) {%
      \includegraphics[width=\qualImageSize]{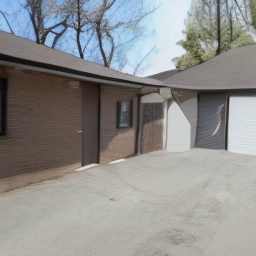}};
    
    \node[myNodeStyle] (img-3-1) at (1,3) {%
      \includegraphics[width=\qualImageSize]{images/qual_results/raw_data/ours/acdcd593a8f50eb1/ground_truth/output_0001.png}};
    \node[myNodeStyle] (img-3-2) at (2,3) {%
      \includegraphics[width=\qualImageSize]{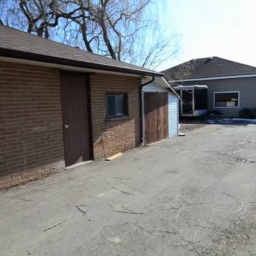}};
    \node[myNodeStyle] (img-3-3) at ({3+\colgap},3) {%
      \includegraphics[width=\qualImageSize]{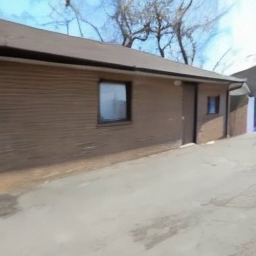}};
    \node[myNodeStyle] (img-3-4) at ({4+\colgap},3) {%
      \includegraphics[width=\qualImageSize]{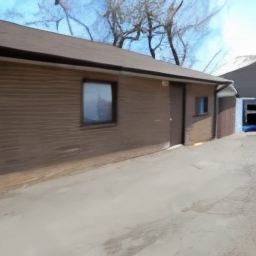}};
    \node[myNodeStyle] (img-3-5) at ({5+\colgap},3) {%
      \includegraphics[width=\qualImageSize]{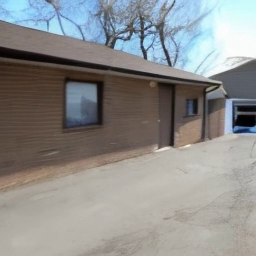}};
    \node[myNodeStyle] (img-3-6) at ({6+\colgap},3) {%
      \includegraphics[width=\qualImageSize]{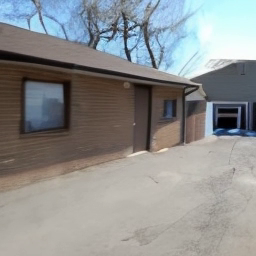}};
    \node[myNodeStyle] (img-3-7) at ({7+\colgap},3) {%
      \includegraphics[width=\qualImageSize]{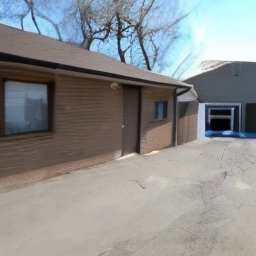}};
    \node[myNodeStyle] (img-3-8) at ({8+\colgap},3) {%
      \includegraphics[width=\qualImageSize]{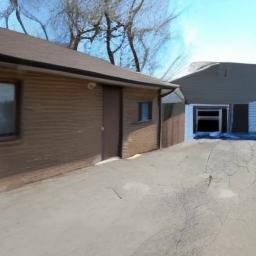}};
    
    \node[myNodeStyle] (img-4-1) at (1,4.1) {%
      \includegraphics[width=\qualImageSize]{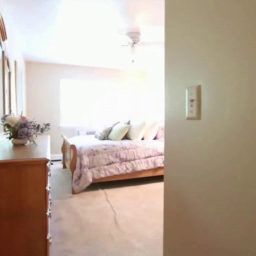}};
    \node[myNodeStyle] (img-4-3) at ({3+\colgap},4.1) {%
      \includegraphics[width=\qualImageSize]{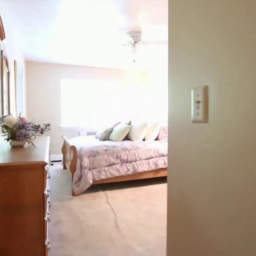}};
    \node[myNodeStyle] (img-4-4) at ({4+\colgap},4.1) {%
      \includegraphics[width=\qualImageSize]{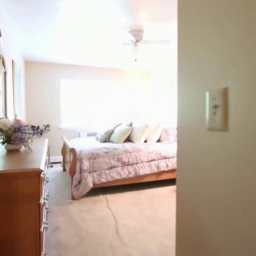}};
    \node[myNodeStyle] (img-4-5) at ({5+\colgap},4.1) {%
      \includegraphics[width=\qualImageSize]{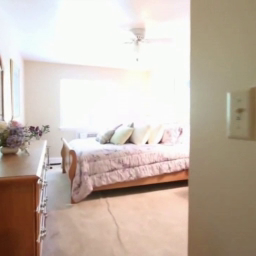}};
    \node[myNodeStyle] (img-4-6) at ({6+\colgap},4.1) {%
      \includegraphics[width=\qualImageSize]{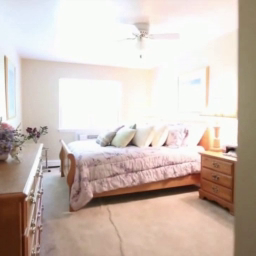}};
    \node[myNodeStyle] (img-4-7) at ({7+\colgap},4.1) {%
      \includegraphics[width=\qualImageSize]{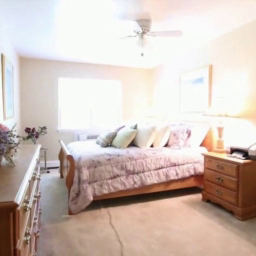}};
    \node[myNodeStyle] (img-4-8) at ({8+\colgap},4.1) {%
      \includegraphics[width=\qualImageSize]{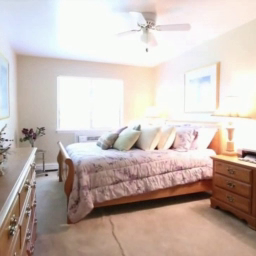}};
    
    \node[myNodeStyle] (img-5-1) at (1,5.1) {%
      \includegraphics[width=\qualImageSize]{images/qual_results/raw_data/ours/01eca393f86d37c5/ground_truth/output_0001.png}};
    \node[myNodeStyle] (img-5-3) at ({3+\colgap},5.1) {%
      \includegraphics[width=\qualImageSize]{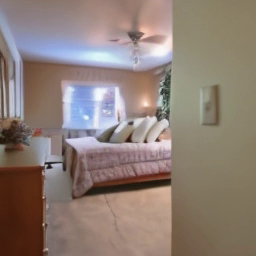}};
    \node[myNodeStyle] (img-5-4) at ({4+\colgap},5.1) {%
      \includegraphics[width=\qualImageSize]{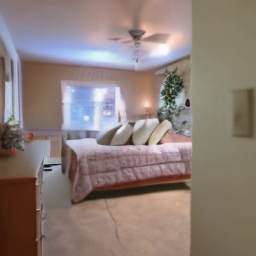}};
    \node[myNodeStyle] (img-5-5) at ({5+\colgap},5.1) {%
      \includegraphics[width=\qualImageSize]{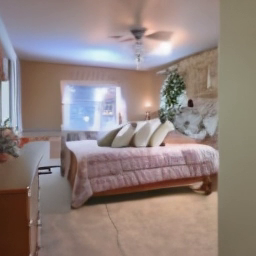}};
    \node[myNodeStyle] (img-5-6) at ({6+\colgap},5.1) {%
      \includegraphics[width=\qualImageSize]{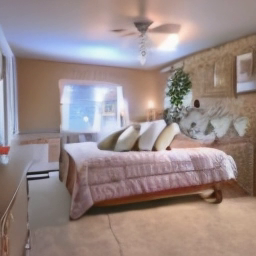}};
    \node[myNodeStyle] (img-5-7) at ({7+\colgap},5.1) {%
      \includegraphics[width=\qualImageSize]{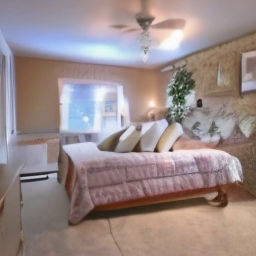}};
    \node[myNodeStyle] (img-5-8) at ({8+\colgap},5.1) {%
      \includegraphics[width=\qualImageSize]{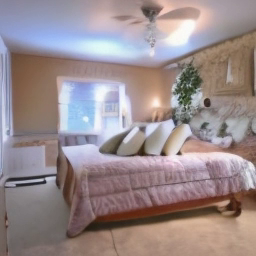}};
    
    \node[myNodeStyle] (img-6-1) at (1,6.1) {%
      \includegraphics[width=\qualImageSize]{images/qual_results/raw_data/ours/01eca393f86d37c5/ground_truth/output_0001.png}};
    \node[myNodeStyle] (img-6-2) at (2,6.1) {%
      \includegraphics[width=\qualImageSize]{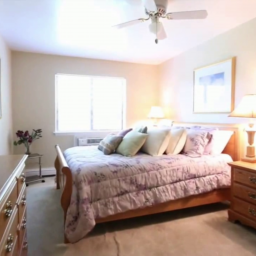}};
    \node[myNodeStyle] (img-6-3) at ({3+\colgap},6.1) {%
      \includegraphics[width=\qualImageSize]{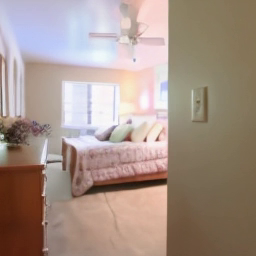}};
    \node[myNodeStyle] (img-6-4) at ({4+\colgap},6.1) {%
      \includegraphics[width=\qualImageSize]{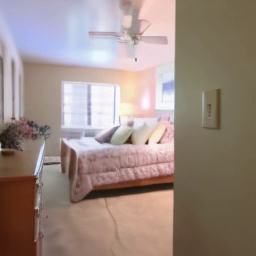}};
    \node[myNodeStyle] (img-6-5) at ({5+\colgap},6.1) {%
      \includegraphics[width=\qualImageSize]{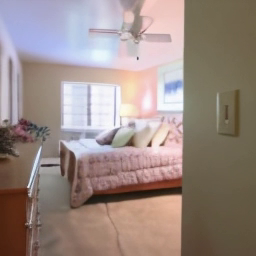}};
    \node[myNodeStyle] (img-6-6) at ({6+\colgap},6.1) {%
      \includegraphics[width=\qualImageSize]{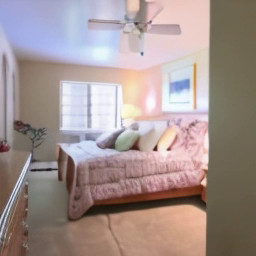}};
    \node[myNodeStyle] (img-6-7) at ({7+\colgap},6.1) {%
      \includegraphics[width=\qualImageSize]{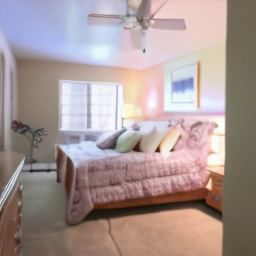}};
    \node[myNodeStyle] (img-6-8) at ({8+\colgap},6.1) {%
      \includegraphics[width=\qualImageSize]{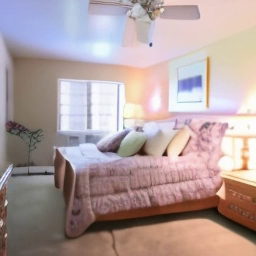}};

\end{tikzpicture}

%% file: figures/qual_result_sampling.tikz
\begin{tikzpicture}[x=1.5cm, y=-1.5cm]
    \newcommand{\qualImageSize}{1.4cm}
    \tikzset{myNodeStyle/.style={draw, inner sep=0pt, outer sep=0pt, minimum size=\qualImageSize}}
    \def\colgap{0.05} 
    \def\headerY{-0.55} 
    
    \node[anchor=south, inner sep=0pt, font=\footnotesize\bfseries] at (1,\headerY) {\textbf{Reference}};
    \node[anchor=south, inner sep=0pt, font=\footnotesize\bfseries] at (2,\headerY) {\textbf{Furthest}};
    \node[anchor=south, inner sep=0pt, font=\footnotesize\bfseries] at (3,\headerY) {\textbf{End+1}};
    
    \node[rotate=90, font=\footnotesize\bfseries] at (0.4,0) {\textbf{GT}};
    \node[rotate=90, font=\footnotesize\bfseries] at (0.4,1) {\textbf{Furthest}};
    \node[rotate=90, font=\footnotesize\bfseries] at (0.4,2) {\textbf{End+1}};
    \node[rotate=90, font=\footnotesize\bfseries] at (0.4,3) {\textbf{Combined}};
    
    \node[myNodeStyle] (box-0-1) at (1,0)           {\includegraphics[width=\qualImageSize]{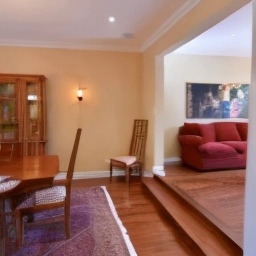}};
    \node[myNodeStyle] (box-0-4) at ({4+\colgap},0) {\includegraphics[width=\qualImageSize]{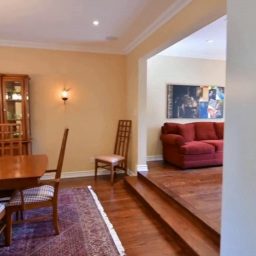}};
    \node[myNodeStyle] (box-0-5) at ({5+\colgap},0) {\includegraphics[width=\qualImageSize]{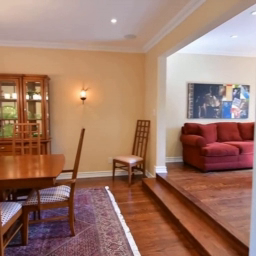}};
    \node[myNodeStyle] (box-0-6) at ({6+\colgap},0) {\includegraphics[width=\qualImageSize]{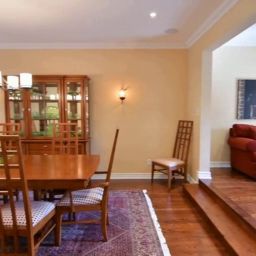}};
    \node[myNodeStyle] (box-0-7) at ({7+\colgap},0) {\includegraphics[width=\qualImageSize]{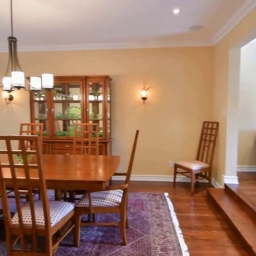}};
    \node[myNodeStyle] (box-0-8) at ({8+\colgap},0) {\includegraphics[width=\qualImageSize]{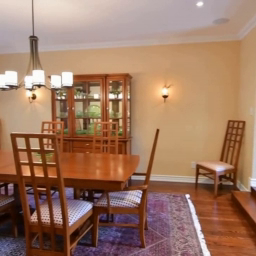}};
    
    \node[myNodeStyle] (box-1-1) at (1,1) {%
      \includegraphics[width=\qualImageSize]{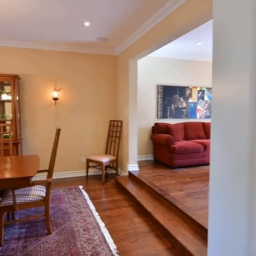}};
    \node[myNodeStyle] (box-1-2) at (2,1) {%
      \includegraphics[width=\qualImageSize]{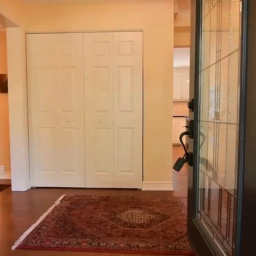}};
    \node[myNodeStyle] (box-1-4) at ({4+\colgap},1) {%
      \includegraphics[width=\qualImageSize]{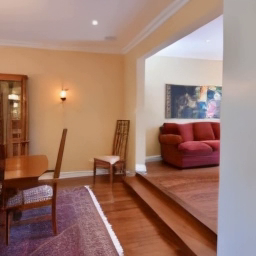}};
    \node[myNodeStyle] (box-1-5) at ({5+\colgap},1) {%
      \includegraphics[width=\qualImageSize]{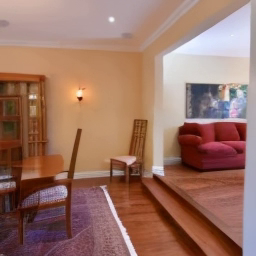}};
    \node[myNodeStyle] (box-1-6) at ({6+\colgap},1) {%
      \includegraphics[width=\qualImageSize]{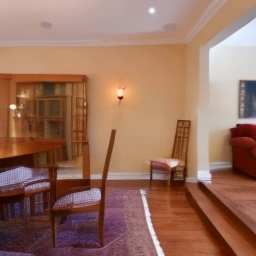}};
    \node[myNodeStyle] (box-1-7) at ({7+\colgap},1) {%
      \includegraphics[width=\qualImageSize]{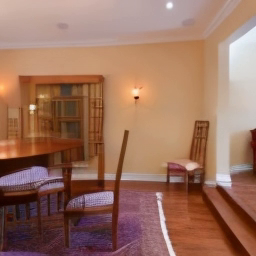}};
    \node[myNodeStyle] (box-1-8) at ({8+\colgap},1) {%
      \includegraphics[width=\qualImageSize]{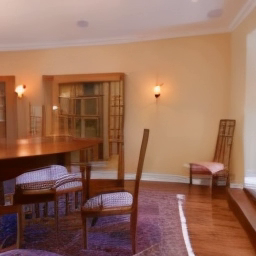}};
    
    \node[myNodeStyle] (box-2-1) at (1,2) {%
      \includegraphics[width=\qualImageSize]{images/qual_results2/904a0f9fa6651f9f_end_furthest/cond_step1.png}};
    \node[myNodeStyle] (box-2-3) at (3,2) {%
      \includegraphics[width=\qualImageSize]{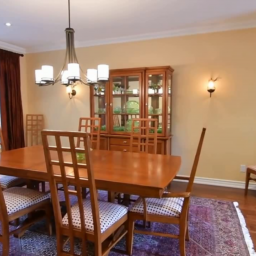}};
    \node[myNodeStyle] (box-2-4) at ({4+\colgap},2) {%
      \includegraphics[width=\qualImageSize]{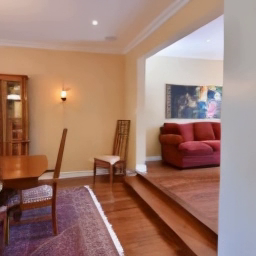}};
    \node[myNodeStyle] (box-2-5) at ({5+\colgap},2) {%
      \includegraphics[width=\qualImageSize]{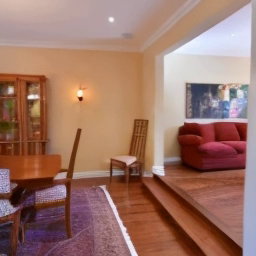}};
    \node[myNodeStyle] (box-2-6) at ({6+\colgap},2) {%
      \includegraphics[width=\qualImageSize]{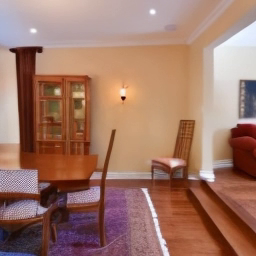}};
    \node[myNodeStyle] (box-2-7) at ({7+\colgap},2) {%
      \includegraphics[width=\qualImageSize]{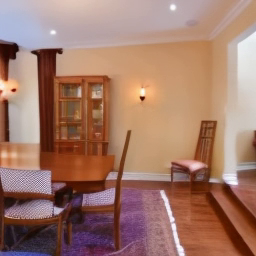}};
    \node[myNodeStyle] (box-2-8) at ({8+\colgap},2) {%
      \includegraphics[width=\qualImageSize]{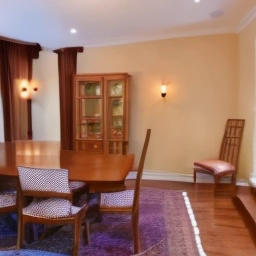}};
    
    \node[myNodeStyle] (box-3-1) at (1,3) {%
      \includegraphics[width=\qualImageSize]{images/qual_results2/904a0f9fa6651f9f_end_furthest/cond_step1.png}};
    \node[myNodeStyle] (box-3-2) at (2,3) {%
      \includegraphics[width=\qualImageSize]{images/qual_results2/904a0f9fa6651f9f_end_furthest/add_cond_step1_1.png}};
    \node[myNodeStyle] (box-3-3) at (3,3) {%
      \includegraphics[width=\qualImageSize]{images/qual_results2/904a0f9fa6651f9f_end_furthest/add_cond_step1_2.png}};
    \node[myNodeStyle] (box-3-4) at ({4+\colgap},3) {%
      \includegraphics[width=\qualImageSize]{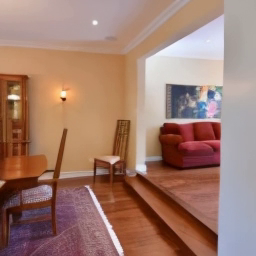}};
    \node[myNodeStyle] (box-3-5) at ({5+\colgap},3) {%
      \includegraphics[width=\qualImageSize]{images/qual_results2/904a0f9fa6651f9f_end_furthest/generated/output_0004}};
    \node[myNodeStyle] (box-3-6) at ({6+\colgap},3) {%
      \includegraphics[width=\qualImageSize]{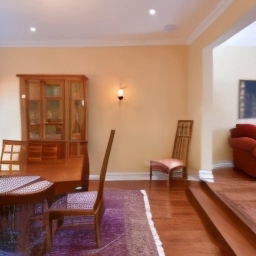}};
    \node[myNodeStyle] (box-3-7) at ({7+\colgap},3) {%
      \includegraphics[width=\qualImageSize]{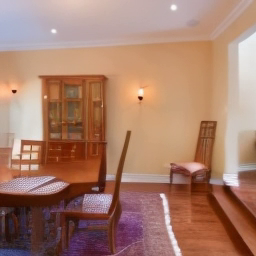}};
    \node[myNodeStyle] (box-3-8) at ({8+\colgap},3) {%
      \includegraphics[width=\qualImageSize]{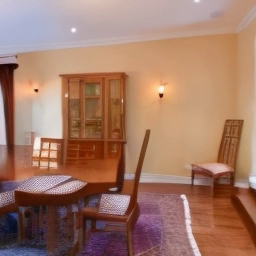}};
\end{tikzpicture}

%% file: sec/5_conclusion.tex
\section{Conclusion and Limitations}\label{sec:conclusion}
This paper introduces CamC2V, a novel conditioning mechanism that provides the diffusion process with extensive contextual information derived from multiple context views.
Unlike conventional image-to-video diffusion models, which typically rely on a single reference image, 
our proposed method employs a \emph{Context-aware Encoder} that encodes additional context through a high-level semantic stream and a 3D-aware visual stream, generating a global semantic representation and a dense, pixel-wise visual embedding from context views. 
This results in significantly improved video fidelity and adherence to scene context, bringing video diffusion models closer to traditional rendering engines.

Still, our method treats the context views as an instantaneous scene snapshot which limits its applicability in highly dynamic scenes.
Moreover, its generative capability is limited by the baseline diffusion model. 
Applying our method on more dynamic scenes and on novel DiT-based diffusion models may be basis to future work. 


%% file: sec/X_suppl.tex
\clearpage
\setcounter{page}{1}
\maketitlesupplementary

\section{Computational Demand}
Table~\ref{tab:comp_demand} reports peak GPU memory and end-to-end inference latency for the baseline models and for our method. As shown, the \emph{Context-aware Encoder} introduces only a small computational overhead relative to the base model. The visual stream exhibits a larger (yet still minor) latency than the semantic stream because it employs additional timestep- and pixel-wise query tokens to produce a dense visual representation.
\begin{table}[h!]
  \centering
  \scriptsize
  \caption{\textbf{Computational demand analysis.} Parameter counts for our method and the baselines, including a decomposition of our \emph{context-aware encoder} into semantic and visual streams. Latency corresponds to the full generation process with 25 DDIM steps under classifier-free guidance (CFG). The encoder adds only minimal overhead and runs comfortably on consumer GPUs with $<\!16$\,GiB of VRAM.}
  \label{tab:comp_demand}
  \begin{tabular}{l
                  rr   
                  c   
                  c} 
    \toprule
    \multirow{2}{*}{Method}
      & \multicolumn{2}{c}{\# Params}
      & \multirow{2}{*}{\makecell{GPU \\ Memory (GiB)}}
      & \multirow{2}{*}{Latency (s)} \\
    \cmidrule(lr){2-3}
      & Trainable & Total
      & 
      &  \\
    \midrule
    DynamiCrafter
      & -  & 2.6\,B
      & 10.43   & 5.143
           \\
    MotionCtrl
      & -  & 2.6\,B
      & 10.49   & 4.849
           \\
   CameraCtrl
      & -  & 2.8\,B
      & 11.28   & 4.898
           \\
   CamI2V
      & -  & 2.9\,B
      & 11.21   &  7.976   \\
      \midrule
    CamC2V
      & 97.4\,M   & 2.9\,B
      & 11.68   & 8.208  \\
     {\quad-- Semantic Stream}
      & 50.9\,M & 50.9\,M
      & 0.19 & 0.004     \\
     {\quad-- Visual Stream}
      & 46.5 \,M  & 46.5\,M 
      &  0.17  & 0.043 \\
    \bottomrule
  \end{tabular}
\end{table}

\section{Camera Evaluation}
We assess the quality of the generated camera trajectories by estimating camera poses with the GLOMAP pipeline. The non-default configuration used in our evaluation is listed in Table~\ref{tab:glomap_params}.

\begin{table}[h!]
  \centering
  \small
  \begin{tabular}{ll}
    \toprule
    \textbf{Parameter} & \textbf{Value} \\
    \midrule
       ImageReader.single\_camera         & 1 \\
       ImageReader.camera\_model          & SIMPLE\_PINHOLE \\
       ImageReader.camera\_params         & \(\{f\},\{cx\},\{cy\}\) \\
       SiftExtraction.estimate\_affine\_shape & 1 \\
       SiftExtraction.domain\_size\_pooling  & 1 \\
    \midrule
       SiftMatching.guided\_matching      & 1 \\
       SiftMatching.max\_num\_matches      & 65536 \\
    \midrule
       RelPoseEstimation.max\_epipolar\_error   & 4 \\
       BundleAdjustment.optimize\_intrinsics   & 0 \\
    \bottomrule
  \end{tabular}
  \caption{\textbf{GLOMAP Configuration.} Changed parameters of the Glomap pipeline used in our evaluation.}
  \label{tab:glomap_params}
\end{table}

\section{Additional Qualitative Results}
To further assess the generative quality we provide further qualitative results of our method in \cref{fig:qual_results} compared to the baseline method CamI2V.
\begin{figure*}[h]
    \centering
    \input{figures/qual_figure4.tikz}
    \caption{\textbf{Additional Qualitative results.} Zoom in for more details}
    \label{fig:qual_results}
\end{figure*}

\section{COLMAP Evaluation}
NeRF or 3D Gaussian Splatting-based approaches often rely on an initial sparse 3D reconstruction step to get an initial pointcloud of the scene.
Typically, COLMAP~\cite{colmap} is used as a sparse 3D reconstruction pipeline. The pipeline can be split in three stages, namely \emph{feature extraction}, \emph{feature matching} and \emph{point triangulation}. After the first stage, we insert the ground-truth camera poses into the pipeline to improve the triangulation performance and kepp it fair to our model, which is also provided with ground-truth camera poses.
\begin{figure}[h]
    \centering
    \input{figures/colmap_success_plot.tikz}
    \caption{\textbf{COLMAP Triangulation.} We report the success rate of sparse two-view reconstruction in COLMAP across increasing inter-frame distances. This serves as the foundation of common NeRF and 3D Gaussian Splatting pipelines. The red dashed line indicates the COLMAP sucess rate for the setup used in our evaluation.}
    \label{fig:colmap_plot}
\end{figure}

\begin{table}[h]
  \centering
  \small
  \begin{tabular}{ll}
    \toprule
    \textbf{Parameter} & \textbf{Value} \\
    \midrule
       ImageReader.single\_camera         & 1 \\
       ImageReader.camera\_model          & SIMPLE\_PINHOLE \\
       ImageReader.camera\_params         & \(\{fx\},\{fy\},\{cx\},\{cy\}\) \\
    \midrule

    \midrule
    Mapper.tri\_ignore\_two\_view\_tracks & 0 \\
    Mapper.num\_threads & 16 \\
    Mapper.init\_min\_tri\_angle & 1 \\
    Mapper.multiple\_models & 0 \\
    Mapper.extract\_colors & 0 \\
    \bottomrule
  \end{tabular}
  \caption{\textbf{COLMAP Configuration.} Changed parameters of the COLMAP pipeline used in our evaluation.}
  \label{tab:glomap_params}
\end{table}

\section{Novel View Synthesis Qualitative Results}
We provide additional qualitative results comparing our model against FrugalNeRF in \cref{fig:qual_results}.
It is visible that FrugalNeRF not only suffers from an insufficient initial sparse reconstruction step, indicated by the black borders, but also introduces artifacts,
especially for fine-grained details and edges, where the 3D representation is not sufficiently learned. 
Overall, our approach mitigates such artifacts and produces clearer more realistic looking images.

\begin{figure*}
    \centering
    \input{figures/frugalnerf_qual_comp.tikz}
    \caption{\textbf{Novel view synthesis qualitative results.} The images show the results from our model compared against the reconstructed scenes from FrugalNeRF. The coniditioning (or training frames for FrugalNeRF) are the first and last image.}
    \label{fig:qual_results}
\end{figure*}

In \cref{fig:colmap_plot}, we show the success rate of this pipeline plotted against the distance between the two used frames.
All failure cases come from the triangulation stage, where the model fails to confidently match and triangulate points between images.
Especially, for larger frame distances the success rate significantly drops, eventhough the images show sufficient overlap in our visual inspection.
This ultimately only leaves a small window from $\sim7$ to $\sim60$ where the pipeline is able to reliably produce a sparse 3D reconstruction in over $80\%$ of the cases. Overall, in our used evaluation setup using the first and $17$th frame the pipeline has a success rate of $83.5$.

%% file: figures/qual_figure4.tikz
\begin{tikzpicture}[x=1.9cm, y=-1.9cm]
    \newcommand{\qualImageSize}{1.8cm}
    \def\colgap{0.05}
    \tikzset{myNodeStyle/.style={draw, inner sep=0pt, outer sep=0pt, minimum size=\qualImageSize}}

    \def\headerY{2.45}
    \node[anchor=south, inner sep=0pt] at (1,\headerY) {\textbf{Reference}};
    \node[anchor=south, inner sep=0pt] at (2,\headerY) {\textbf{Context}};

    \node[rotate=90] at (0.4,3) {\textbf{GT}};
    \node[rotate=90] at (0.4,4) {\textbf{CamI2V}};
    \node[rotate=90] at (0.4,5) {\textbf{Ours}};

    \node[myNodeStyle] at (1,3) {%
      \includegraphics[width=\qualImageSize]{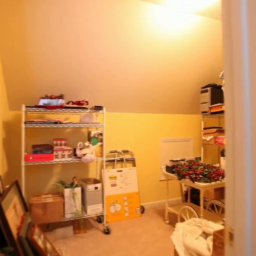}};
    \node[myNodeStyle] at ({3+\colgap},3) {%
      \includegraphics[width=\qualImageSize]{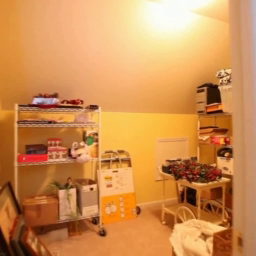}};
    \node[myNodeStyle] at ({4+\colgap},3) {%
      \includegraphics[width=\qualImageSize]{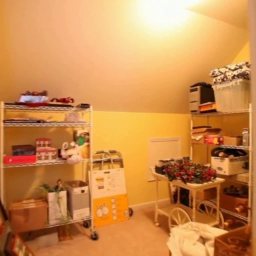}};
    \node[myNodeStyle] at ({5+\colgap},3) {%
      \includegraphics[width=\qualImageSize]{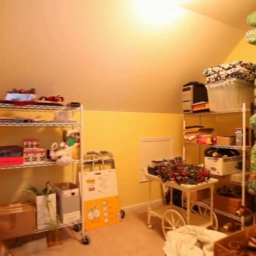}};
    \node[myNodeStyle] at ({6+\colgap},3) {%
      \includegraphics[width=\qualImageSize]{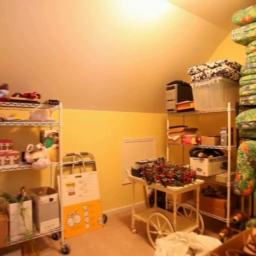}};
    \node[myNodeStyle] at ({7+\colgap},3) {%
      \includegraphics[width=\qualImageSize]{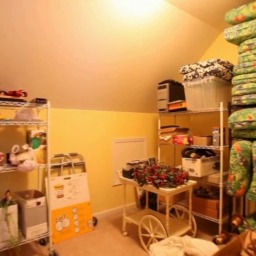}};
    \node[myNodeStyle] at ({8+\colgap},3) {%
      \includegraphics[width=\qualImageSize]{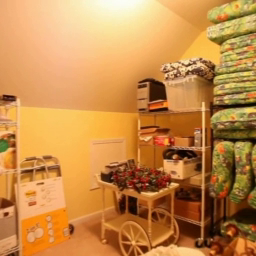}};

    \node[myNodeStyle] at (1,4) {%
      \includegraphics[width=\qualImageSize]{images/add_qual_results/022a21a897f2a904_2/cond_step1.png}};
    \node[myNodeStyle] at ({3+\colgap},4) {%
      \includegraphics[width=\qualImageSize]{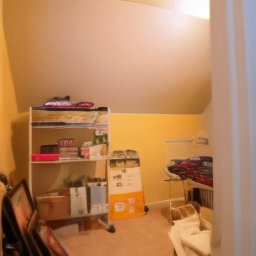}};
    \node[myNodeStyle] at ({4+\colgap},4) {%
      \includegraphics[width=\qualImageSize]{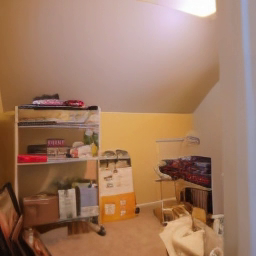}};
    \node[myNodeStyle] at ({5+\colgap},4) {%
      \includegraphics[width=\qualImageSize]{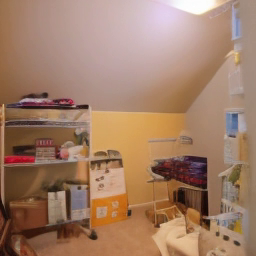}};
    \node[myNodeStyle] at ({6+\colgap},4) {%
      \includegraphics[width=\qualImageSize]{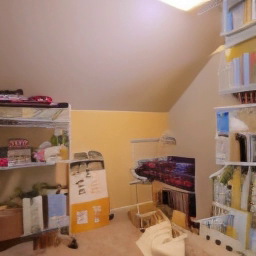}};
    \node[myNodeStyle] at ({7+\colgap},4) {%
      \includegraphics[width=\qualImageSize]{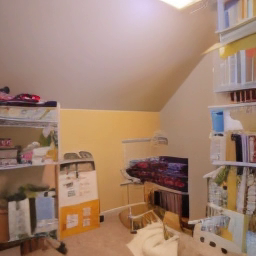}};
    \node[myNodeStyle] at ({8+\colgap},4) {%
      \includegraphics[width=\qualImageSize]{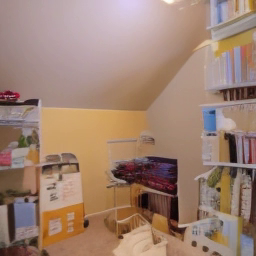}};

    \node[myNodeStyle] at (1,5) {%
      \includegraphics[width=\qualImageSize]{images/add_qual_results/022a21a897f2a904_2/cond_step1.png}};
    \node[myNodeStyle] at (2,5) {%
      \includegraphics[width=\qualImageSize]{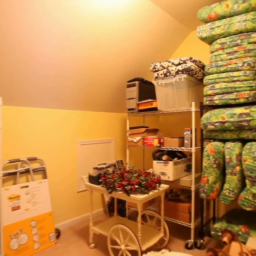}};
    \node[myNodeStyle] at ({3+\colgap},5) {%
      \includegraphics[width=\qualImageSize]{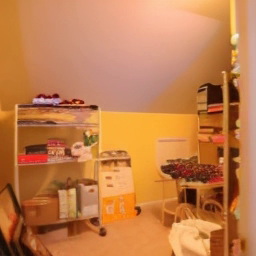}};
    \node[myNodeStyle] at ({4+\colgap},5) {%
      \includegraphics[width=\qualImageSize]{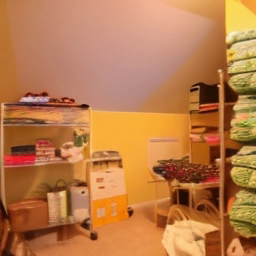}};
    \node[myNodeStyle] at ({5+\colgap},5) {%
      \includegraphics[width=\qualImageSize]{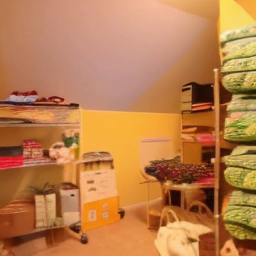}};
    \node[myNodeStyle] at ({6+\colgap},5) {%
      \includegraphics[width=\qualImageSize]{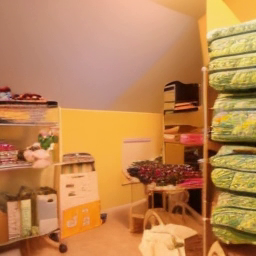}};
    \node[myNodeStyle] at ({7+\colgap},5) {%
      \includegraphics[width=\qualImageSize]{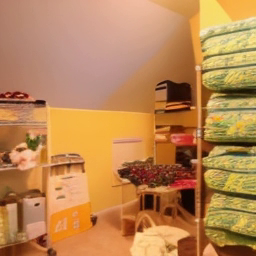}};
    \node[myNodeStyle] at ({8+\colgap},5) {%
      \includegraphics[width=\qualImageSize]{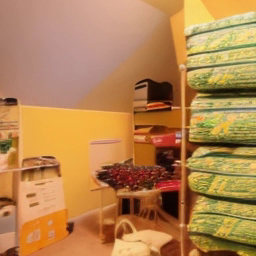}};


    \node[rotate=90] at (0.4,6.1) {\textbf{GT}};
    \node[myNodeStyle] at (1,6.1) {%
      \includegraphics[width=\qualImageSize]{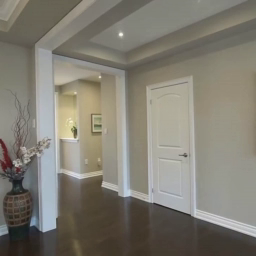}};
    \node[myNodeStyle] at ({3+\colgap},6.1) {%
      \includegraphics[width=\qualImageSize]{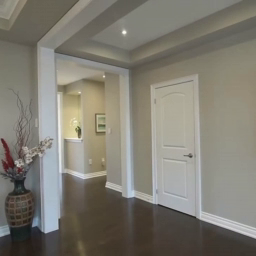}};
    \node[myNodeStyle] at ({4+\colgap},6.1) {%
      \includegraphics[width=\qualImageSize]{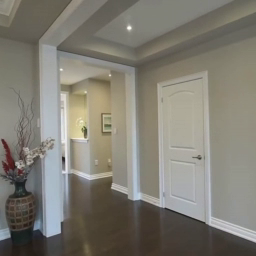}};
    \node[myNodeStyle] at ({5+\colgap},6.1) {%
      \includegraphics[width=\qualImageSize]{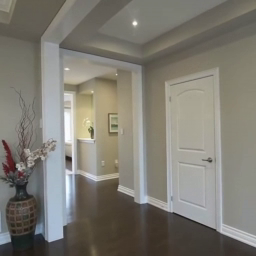}};
    \node[myNodeStyle] at ({6+\colgap},6.1) {%
      \includegraphics[width=\qualImageSize]{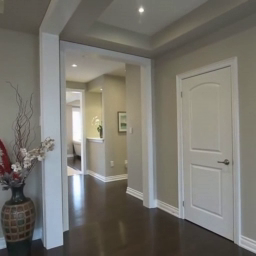}};
    \node[myNodeStyle] at ({7+\colgap},6.1) {%
      \includegraphics[width=\qualImageSize]{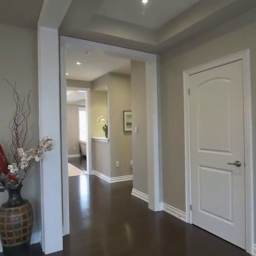}};
    \node[myNodeStyle] at ({8+\colgap},6.1) {%
      \includegraphics[width=\qualImageSize]{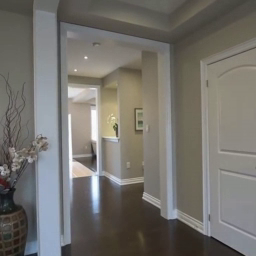}};

    \node[rotate=90] at (0.4,7.1) {\textbf{CamI2V}};
    \node[myNodeStyle] at (1,7.1) {%
      \includegraphics[width=\qualImageSize]{images/qual_results/raw_data/ours/ea238737feca4622/ground_truth/output_0001.png}};
    \node[myNodeStyle] at ({3+\colgap},7.1) {%
      \includegraphics[width=\qualImageSize]{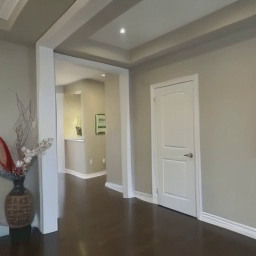}};
    \node[myNodeStyle] at ({4+\colgap},7.1) {%
      \includegraphics[width=\qualImageSize]{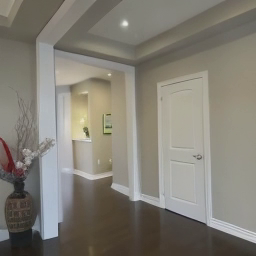}};
    \node[myNodeStyle] at ({5+\colgap},7.1) {%
      \includegraphics[width=\qualImageSize]{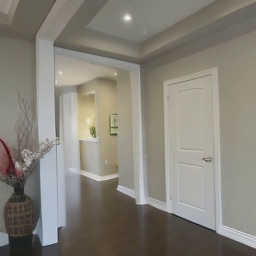}};
    \node[myNodeStyle] at ({6+\colgap},7.1) {%
      \includegraphics[width=\qualImageSize]{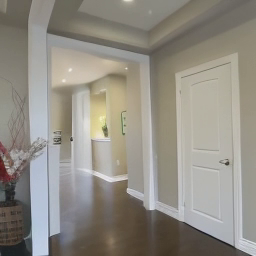}};
    \node[myNodeStyle] at ({7+\colgap},7.1) {%
      \includegraphics[width=\qualImageSize]{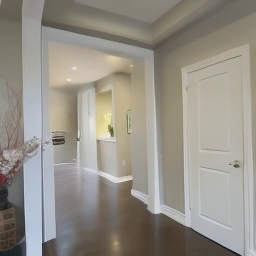}};
    \node[myNodeStyle] at ({8+\colgap},7.1) {%
      \includegraphics[width=\qualImageSize]{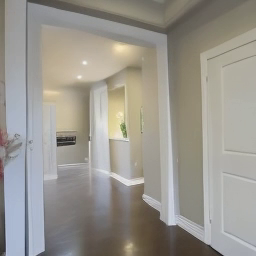}};

    \node[rotate=90] at (0.4,8.1) {\textbf{Ours}};
    \node[myNodeStyle] at (1,8.1) {%
      \includegraphics[width=\qualImageSize]{images/qual_results/raw_data/ours/ea238737feca4622/ground_truth/output_0001.png}};
    \node[myNodeStyle] at (2,8.1) {%
      \includegraphics[width=\qualImageSize]{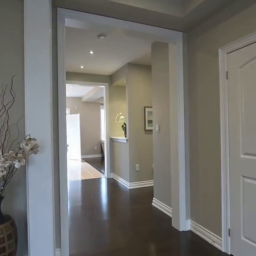}};
    \node[myNodeStyle] at ({3+\colgap},8.1) {%
      \includegraphics[width=\qualImageSize]{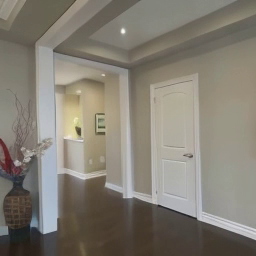}};
    \node[myNodeStyle] at ({4+\colgap},8.1) {%
      \includegraphics[width=\qualImageSize]{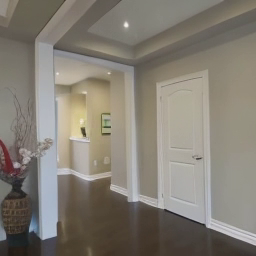}};
    \node[myNodeStyle] at ({5+\colgap},8.1) {%
      \includegraphics[width=\qualImageSize]{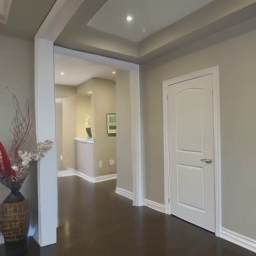}};
    \node[myNodeStyle] at ({6+\colgap},8.1) {%
      \includegraphics[width=\qualImageSize]{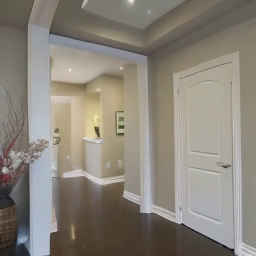}};
    \node[myNodeStyle] at ({7+\colgap},8.1) {%
      \includegraphics[width=\qualImageSize]{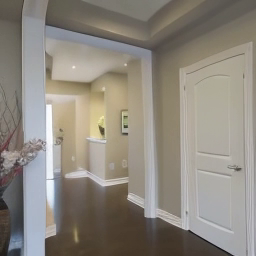}};
    \node[myNodeStyle] at ({8+\colgap},8.1) {%
      \includegraphics[width=\qualImageSize]{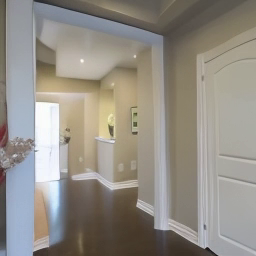}};

\end{tikzpicture}

%% file: figures/colmap_success_plot.tikz
\begin{tikzpicture}
\pgfmathsetmacro{\avg}{0.51223}
\begin{axis}[
    width=9cm, height=4.5cm,
    title={COLMAP Success Rate},
    xlabel={Frame Distance},
    xmin=0, xmax=200,
    ymin=0, ymax=1.0,
    xtick={0,25,50,75,100,125,150,175,200},
    ytick={0,0.2,0.4,0.6,0.8,1.0},
    grid=both,
    grid style={densely dotted},
    tick align=outside,
    tick style={thick},
    enlargelimits=false,
]

\addplot+[
    very thick,
    no marks,
    smooth,
    color=blue!70!cyan
] coordinates {
    (0,0.0)
    (1,0.025)
    (3,0.345)
    (5,0.65)
    (10,0.82)
    (16,0.835)
    (20,0.84)
    (50,0.874)
    (75,0.655)
    (100,0.57)
    (150,0.57)
    (200,0.275)
    (250,0.2)
};

\addplot [dashed, thick, domain=0:200, samples=2, no marks, color=black!60] {\avg};
\node[
    anchor=south east,
    fill=white,
    draw=black!30,
    rounded corners=2pt,
    inner sep=3pt,
    xshift=-2pt, yshift=2pt,
    font=\footnotesize
] at (axis cs:198,\avg+0.1) {Average: \pgfmathprintnumber[fixed,precision=3]{\avg}};

\addplot[densely dashed, very thick, red!70] coordinates {(16,0) (16,0.835)};

\node[
    anchor=west,
    fill=white,
    draw=red!70,
    rounded corners=2pt,
    inner sep=3pt,
    font=\footnotesize
] at (axis cs:20,0.4) {Success rate: 0.835};

\node[
    anchor=north,
    font=\footnotesize\bfseries,
    text=red!70
] at (axis cs:16,0) {16};

\end{axis}
\end{tikzpicture}

%% file: figures/frugalnerf_qual_comp.tikz
\begin{tikzpicture}[x=1.9cm, y=-1.9cm]
    \newcommand{\qualImageSize}{1.8cm}
    \def\colgap{0.0}
    \tikzset{myNodeStyle/.style={draw, inner sep=0pt, outer sep=0pt, minimum size=\qualImageSize}}

    \def\headerY{2.45}

    \node[rotate=90] at (0.4,3) {\textbf{GT}};
    \node[rotate=90] at (0.4,4) {\textbf{FrugalNeRF}};
    \node[rotate=90] at (0.4,5) {\textbf{Ours}};

    \node[myNodeStyle] at ({1+\colgap},3) {\includegraphics[width=\qualImageSize]{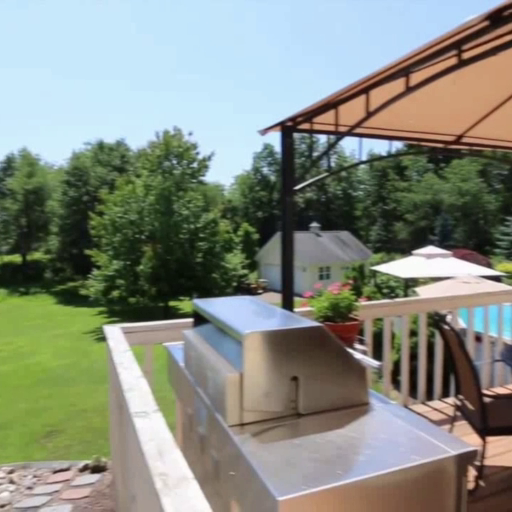}};
    \node[myNodeStyle] at ({2+\colgap},3) {\includegraphics[width=\qualImageSize]{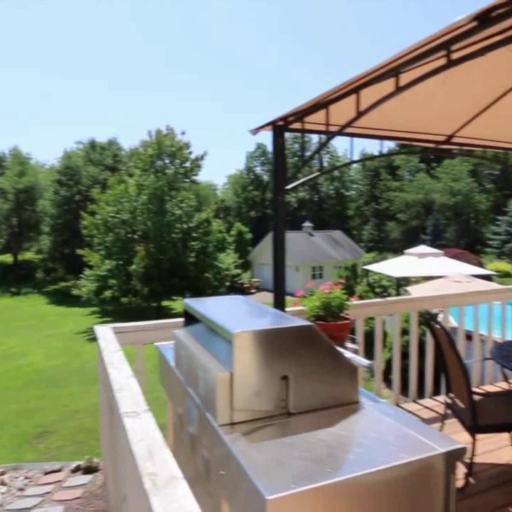}};
    \node[myNodeStyle] at ({3+\colgap},3) {\includegraphics[width=\qualImageSize]{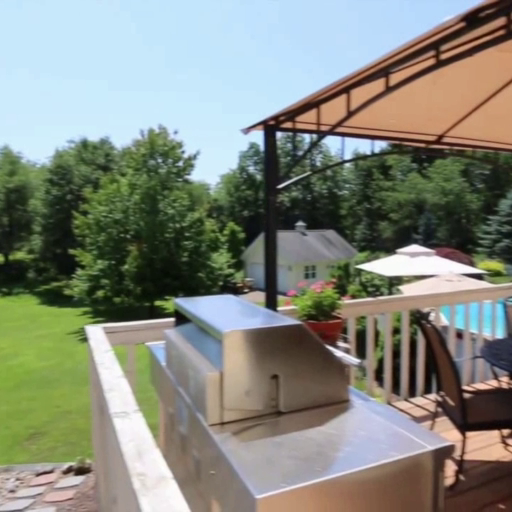}};
    \node[myNodeStyle] at ({4+\colgap},3) {\includegraphics[width=\qualImageSize]{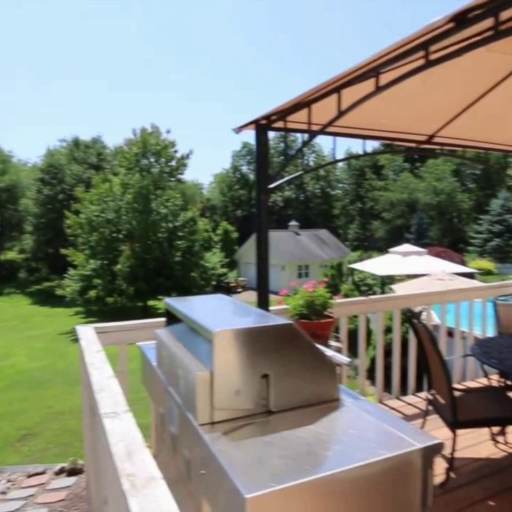}};
    \node[myNodeStyle] at ({5+\colgap},3) {\includegraphics[width=\qualImageSize]{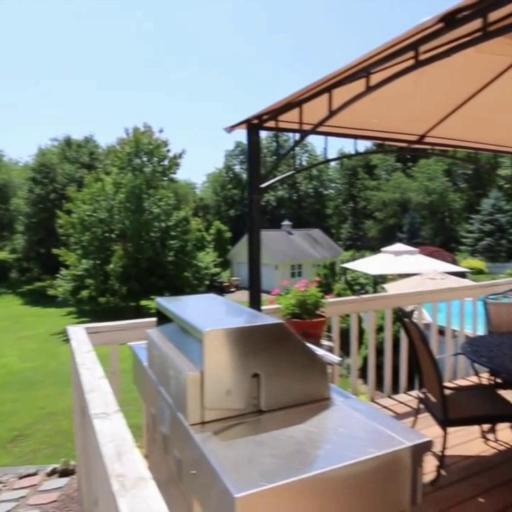}};
    \node[myNodeStyle] at ({6+\colgap},3) {\includegraphics[width=\qualImageSize]{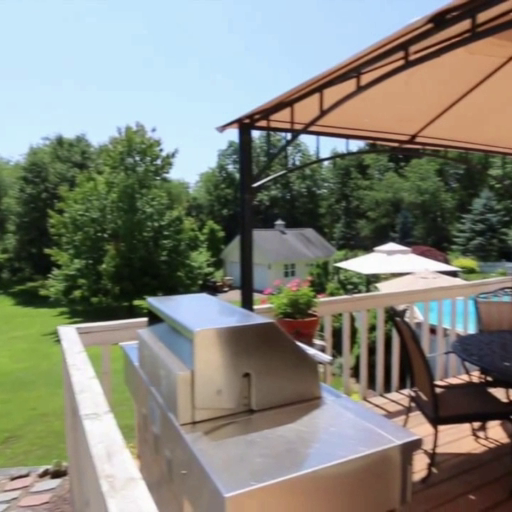}};
    \node[myNodeStyle] at ({7+\colgap},3) {\includegraphics[width=\qualImageSize]{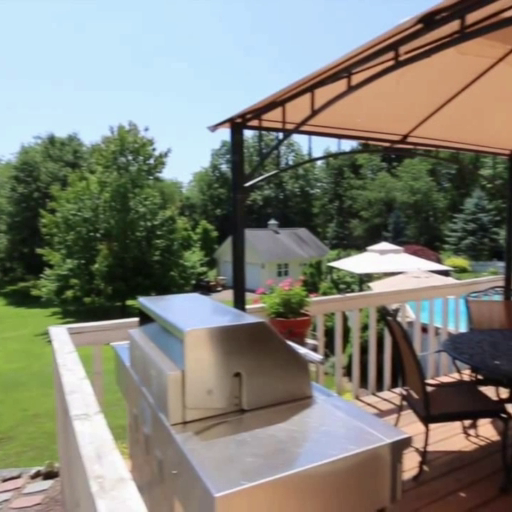}};
    \node[myNodeStyle] at ({8+\colgap},3) {\includegraphics[width=\qualImageSize]{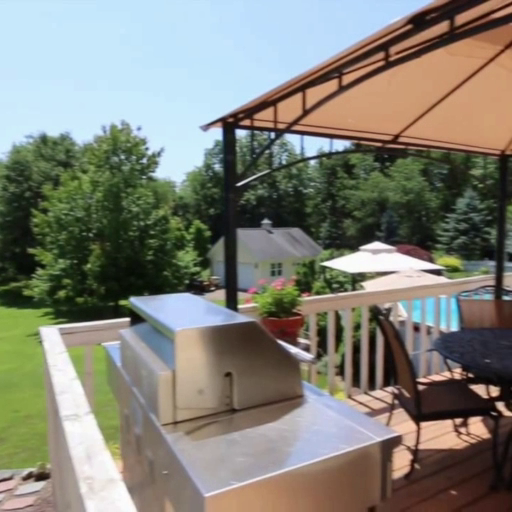}};

    \node[myNodeStyle] at ({1+\colgap},4) {\includegraphics[width=\qualImageSize]{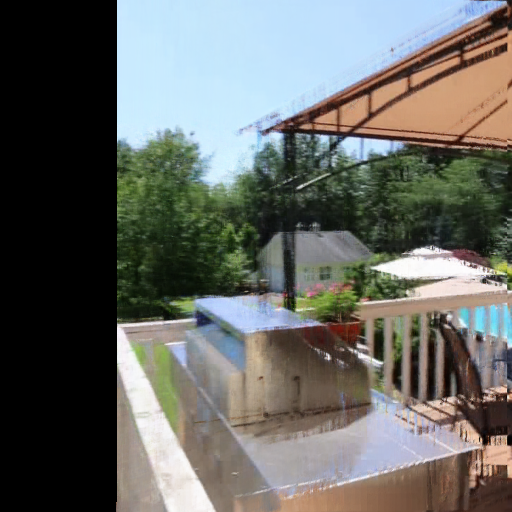}};
    \node[myNodeStyle] at ({2+\colgap},4) {\includegraphics[width=\qualImageSize]{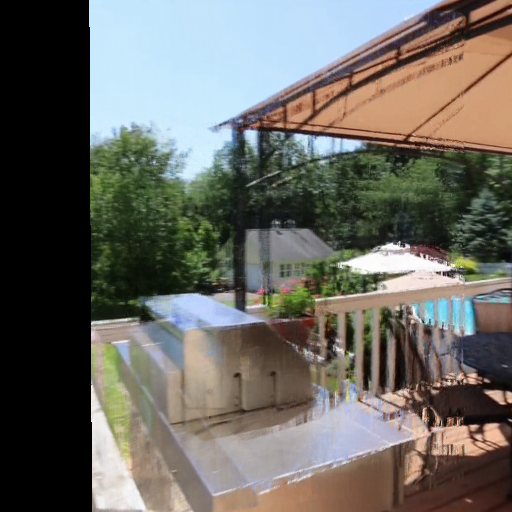}};
    \node[myNodeStyle] at ({3+\colgap},4) {\includegraphics[width=\qualImageSize]{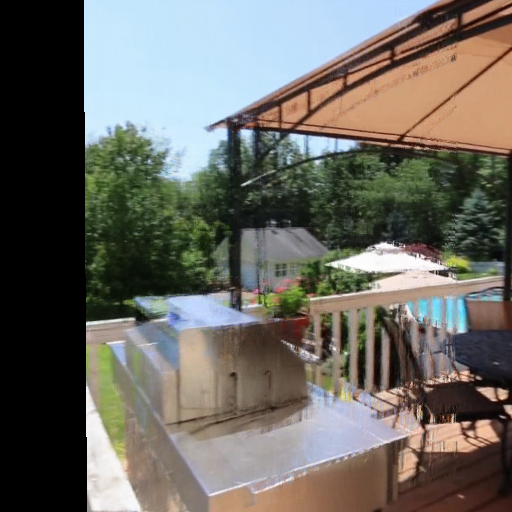}};
    \node[myNodeStyle] at ({4+\colgap},4) {\includegraphics[width=\qualImageSize]{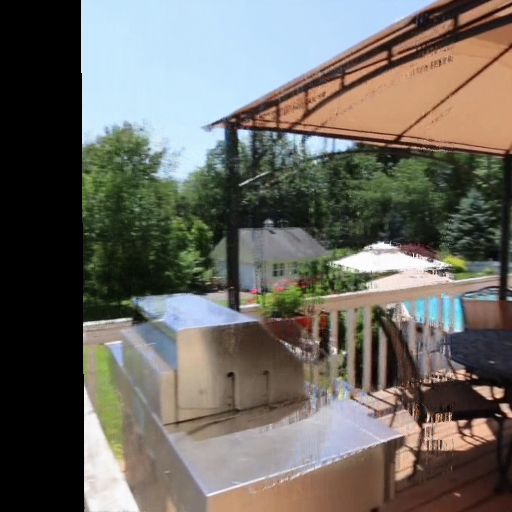}};
    \node[myNodeStyle] at ({5+\colgap},4) {\includegraphics[width=\qualImageSize]{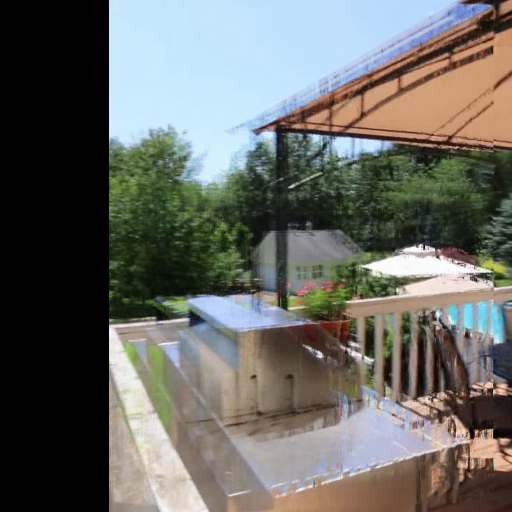}};
    \node[myNodeStyle] at ({6+\colgap},4) {\includegraphics[width=\qualImageSize]{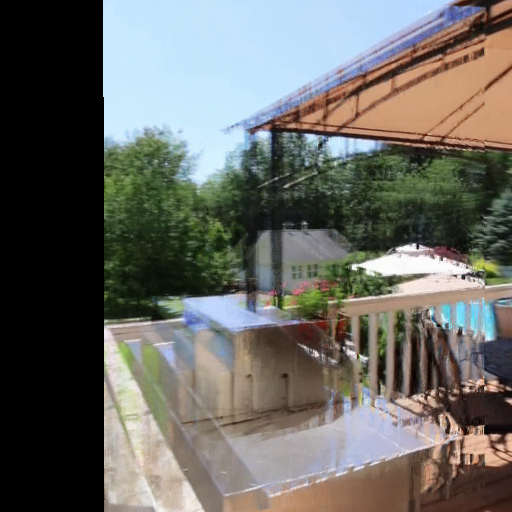}};
    \node[myNodeStyle] at ({7+\colgap},4) {\includegraphics[width=\qualImageSize]{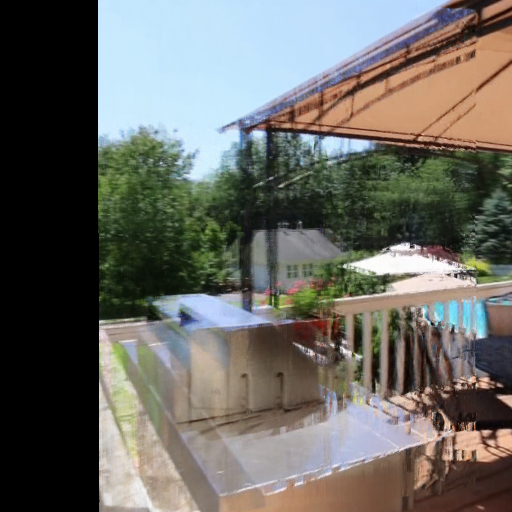}};
    \node[myNodeStyle] at ({8+\colgap},4) {\includegraphics[width=\qualImageSize]{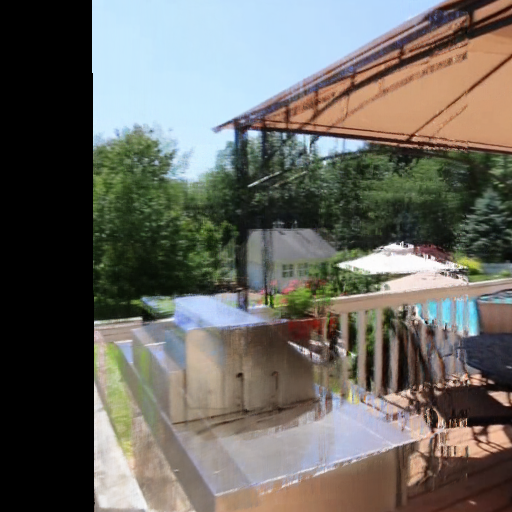}};

    \node[myNodeStyle] at ({1+\colgap},5) {\includegraphics[width=\qualImageSize]{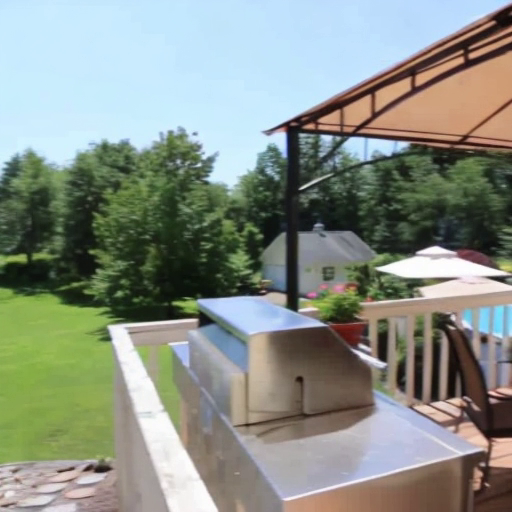}};
    \node[myNodeStyle] at ({2+\colgap},5) {\includegraphics[width=\qualImageSize]{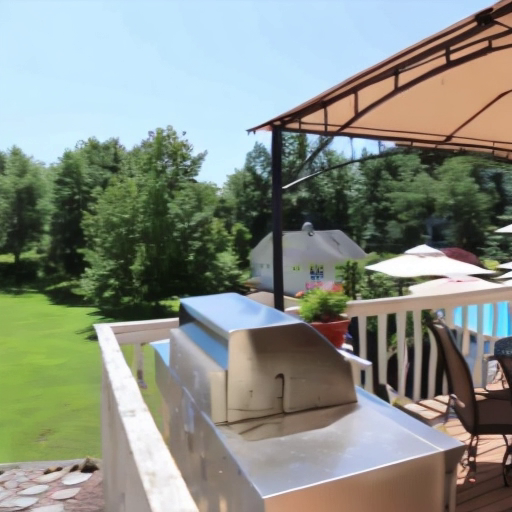}};
    \node[myNodeStyle] at ({3+\colgap},5) {\includegraphics[width=\qualImageSize]{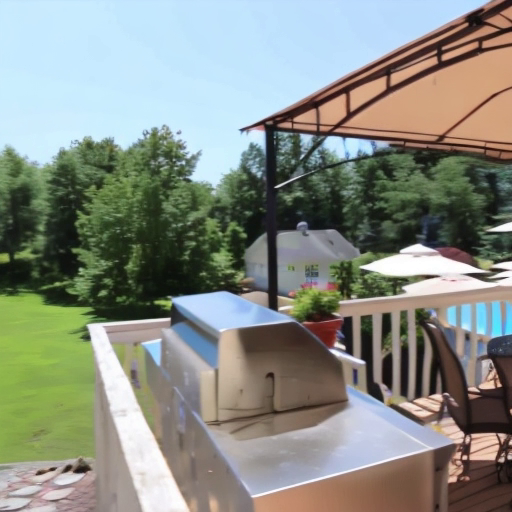}};
    \node[myNodeStyle] at ({4+\colgap},5) {\includegraphics[width=\qualImageSize]{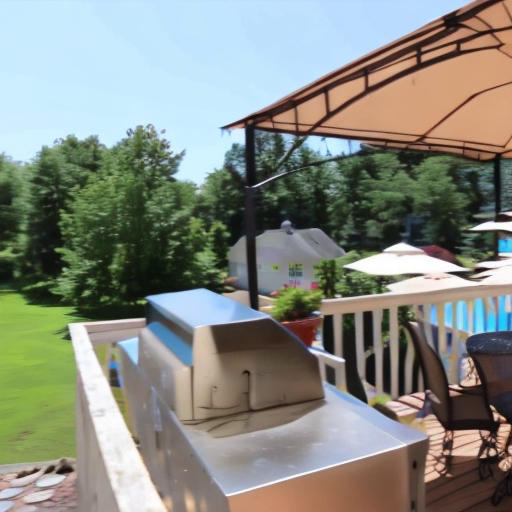}};
    \node[myNodeStyle] at ({5+\colgap},5) {\includegraphics[width=\qualImageSize]{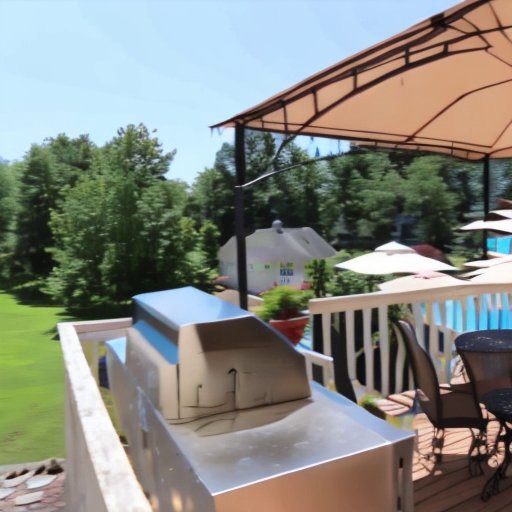}};
    \node[myNodeStyle] at ({6+\colgap},5) {\includegraphics[width=\qualImageSize]{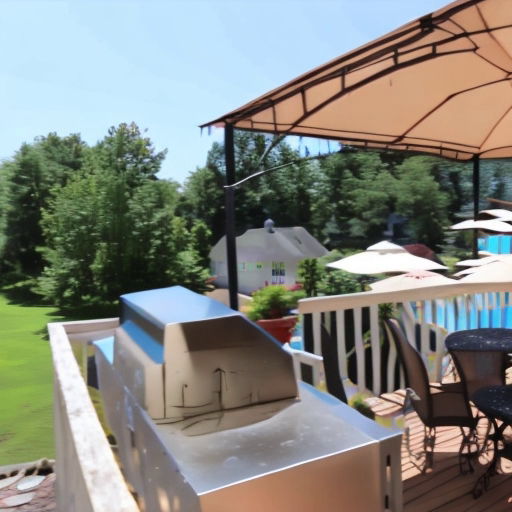}};
    \node[myNodeStyle] at ({7+\colgap},5) {\includegraphics[width=\qualImageSize]{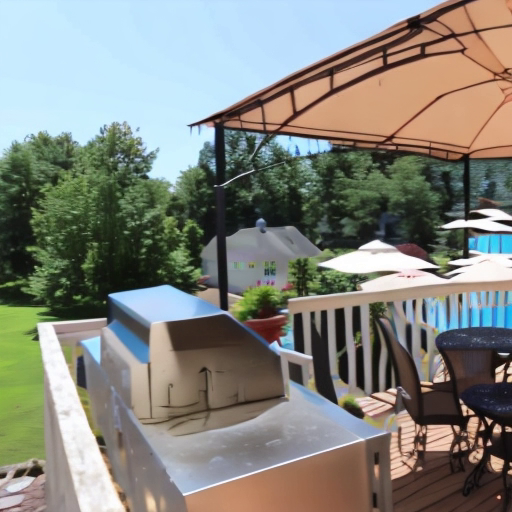}};
    \node[myNodeStyle] at ({8+\colgap},5) {\includegraphics[width=\qualImageSize]{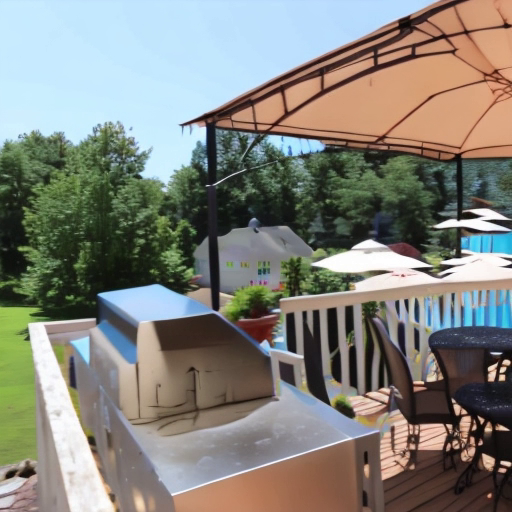}};

    \node[rotate=90] at (0.4,6.1) {\textbf{GT}};
    \node[myNodeStyle] at ({1+\colgap},6.1) {\includegraphics[width=\qualImageSize]{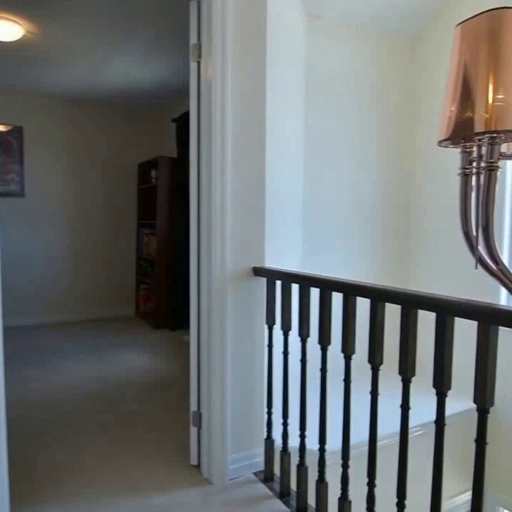}};
    \node[myNodeStyle] at ({2+\colgap},6.1) {\includegraphics[width=\qualImageSize]{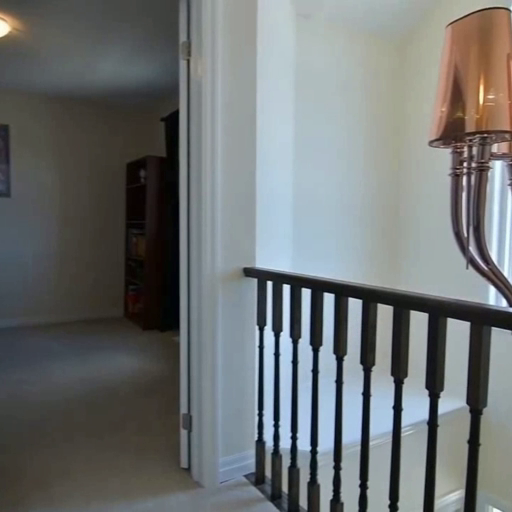}};
    \node[myNodeStyle] at ({3+\colgap},6.1) {\includegraphics[width=\qualImageSize]{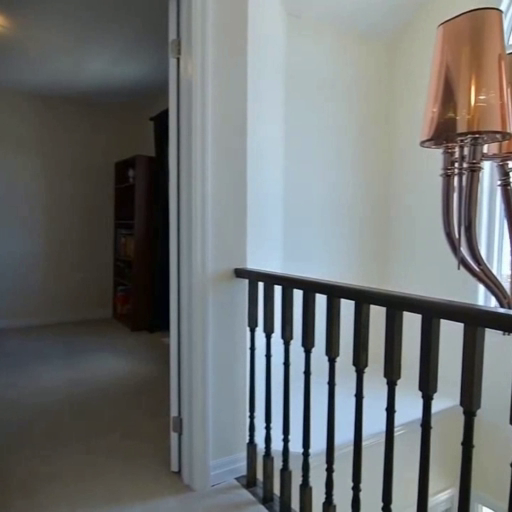}};
    \node[myNodeStyle] at ({4+\colgap},6.1) {\includegraphics[width=\qualImageSize]{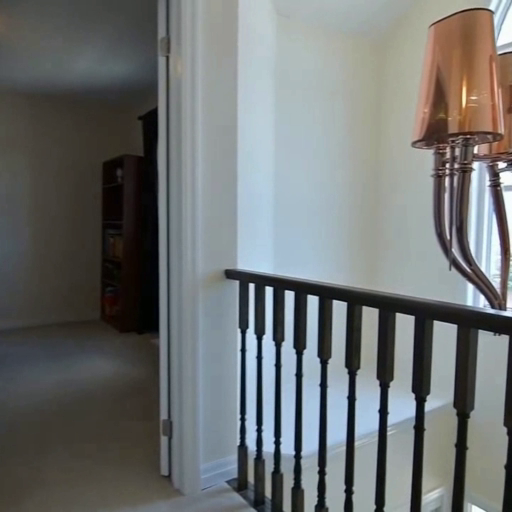}};
    \node[myNodeStyle] at ({5+\colgap},6.1) {\includegraphics[width=\qualImageSize]{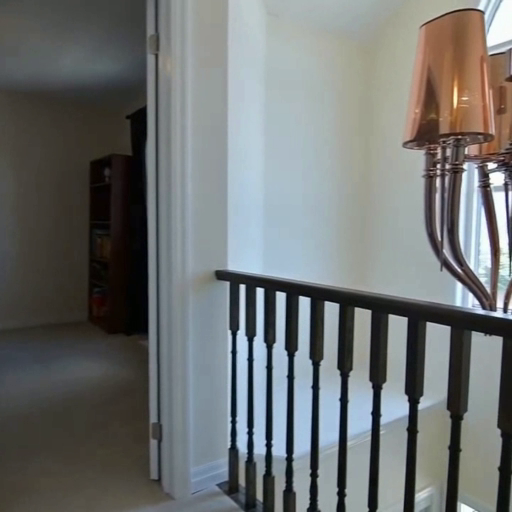}};
    \node[myNodeStyle] at ({6+\colgap},6.1) {\includegraphics[width=\qualImageSize]{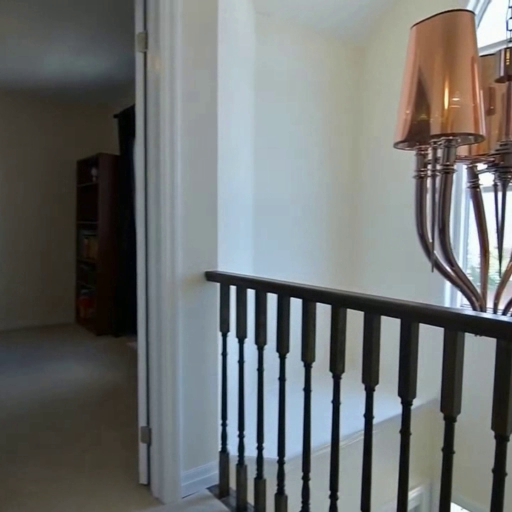}};
    \node[myNodeStyle] at ({7+\colgap},6.1) {\includegraphics[width=\qualImageSize]{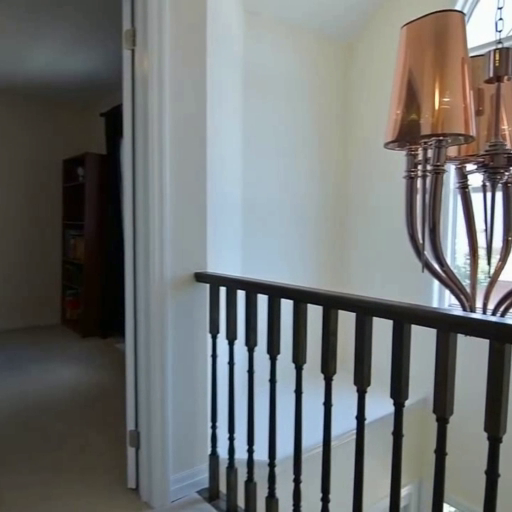}};
    \node[myNodeStyle] at ({8+\colgap},6.1) {\includegraphics[width=\qualImageSize]{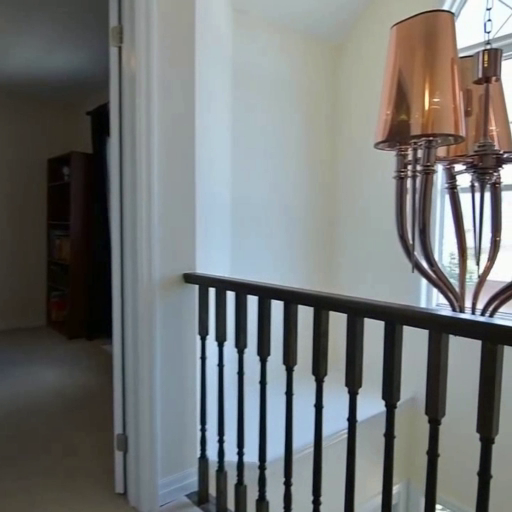}};

    \node[rotate=90] at (0.4,7.1) {\textbf{FrugalNeRF}};
    \node[myNodeStyle] at ({1+\colgap},7.1) {\includegraphics[width=\qualImageSize]{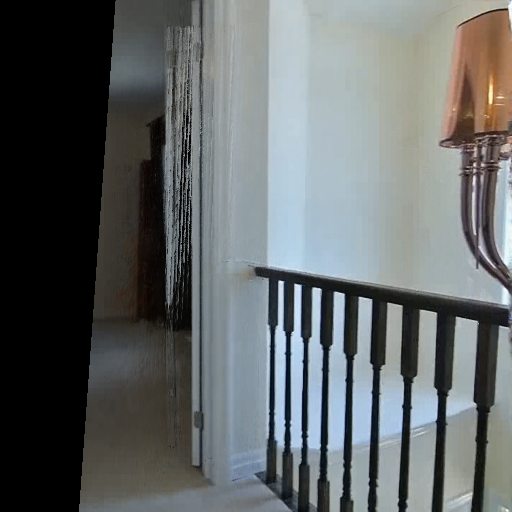}};
    \node[myNodeStyle] at ({2+\colgap},7.1) {\includegraphics[width=\qualImageSize]{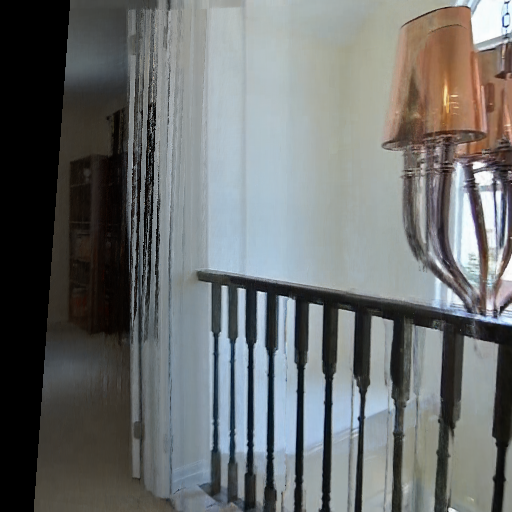}};
    \node[myNodeStyle] at ({3+\colgap},7.1) {\includegraphics[width=\qualImageSize]{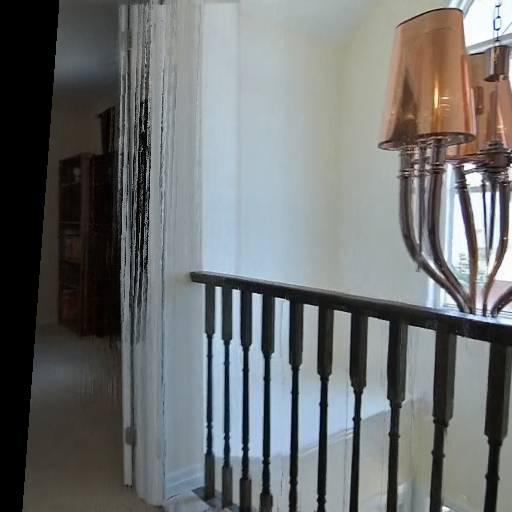}};
    \node[myNodeStyle] at ({4+\colgap},7.1) {\includegraphics[width=\qualImageSize]{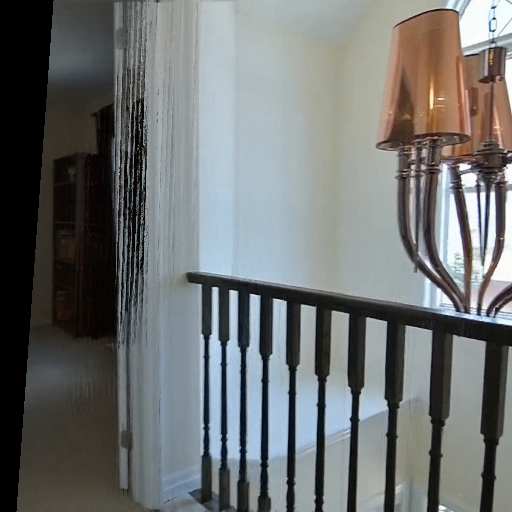}};
    \node[myNodeStyle] at ({5+\colgap},7.1) {\includegraphics[width=\qualImageSize]{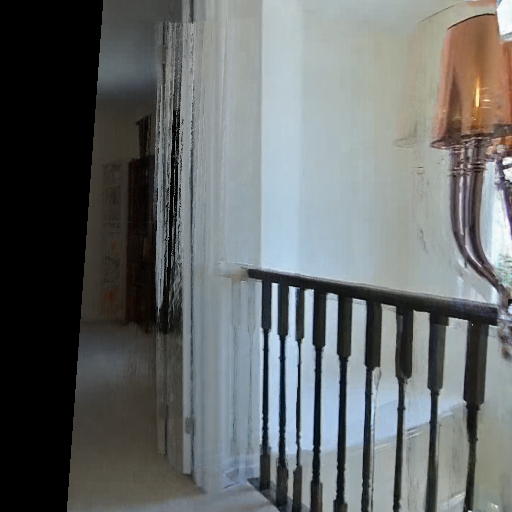}};
    \node[myNodeStyle] at ({6+\colgap},7.1) {\includegraphics[width=\qualImageSize]{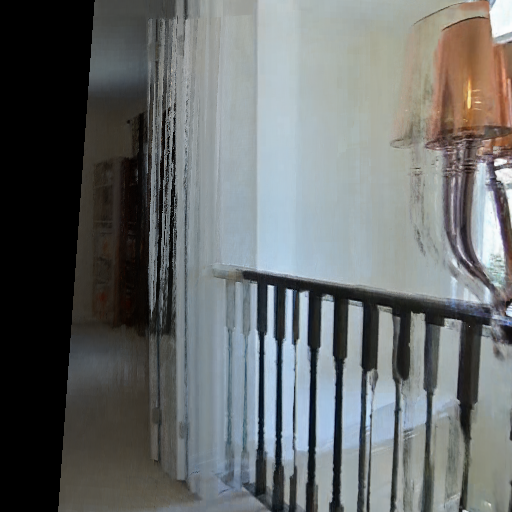}};
    \node[myNodeStyle] at ({7+\colgap},7.1) {\includegraphics[width=\qualImageSize]{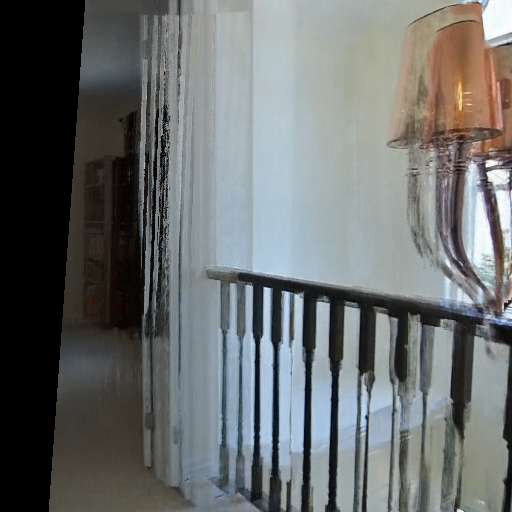}};
    \node[myNodeStyle] at ({8+\colgap},7.1) {\includegraphics[width=\qualImageSize]{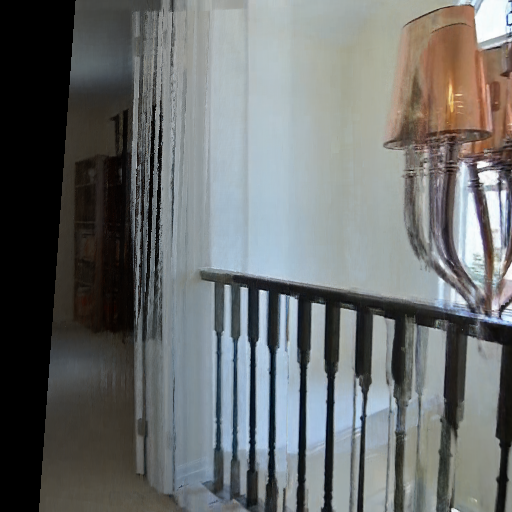}};

    \node[rotate=90] at (0.4,8.1) {\textbf{Ours}};
    \node[myNodeStyle] at ({1+\colgap},8.1) {\includegraphics[width=\qualImageSize]{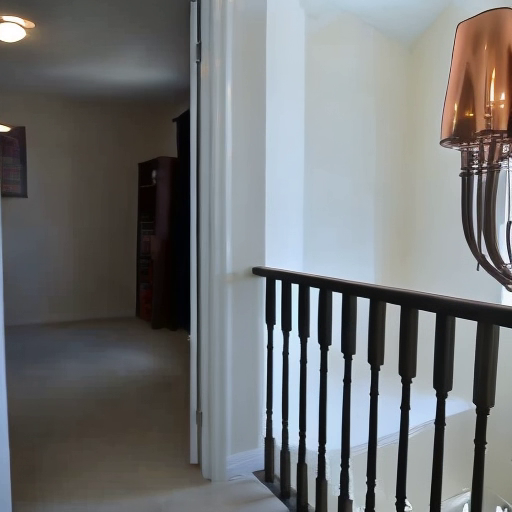}};
    \node[myNodeStyle] at ({2+\colgap},8.1) {\includegraphics[width=\qualImageSize]{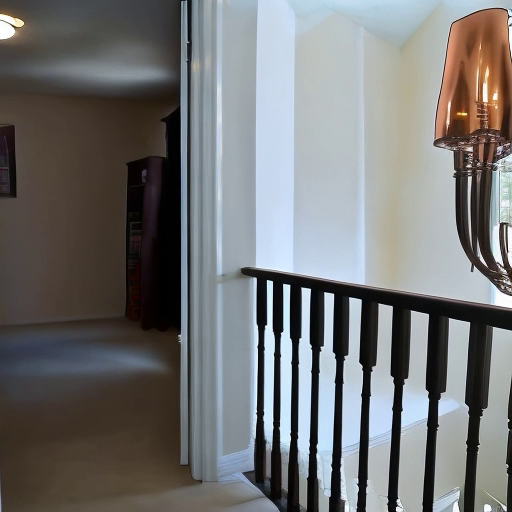}};
    \node[myNodeStyle] at ({3+\colgap},8.1) {\includegraphics[width=\qualImageSize]{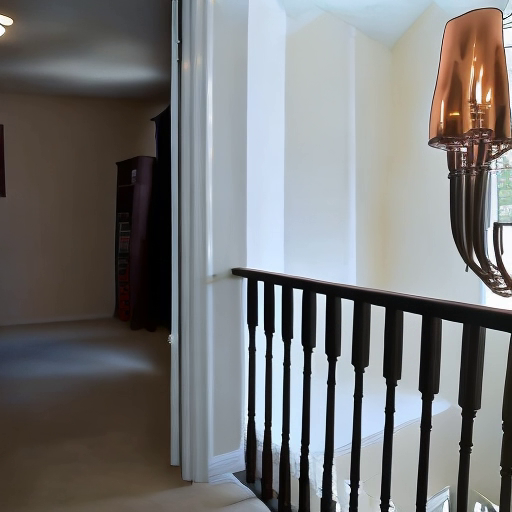}};
    \node[myNodeStyle] at ({4+\colgap},8.1) {\includegraphics[width=\qualImageSize]{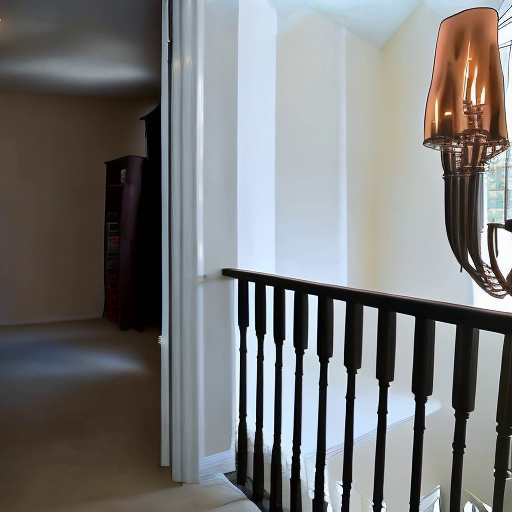}};
    \node[myNodeStyle] at ({5+\colgap},8.1) {\includegraphics[width=\qualImageSize]{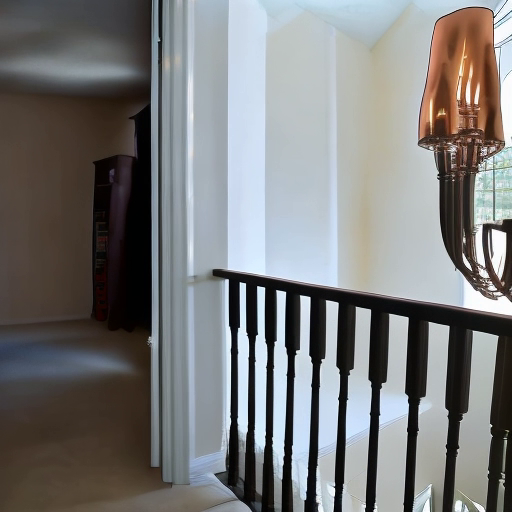}};
    \node[myNodeStyle] at ({6+\colgap},8.1) {\includegraphics[width=\qualImageSize]{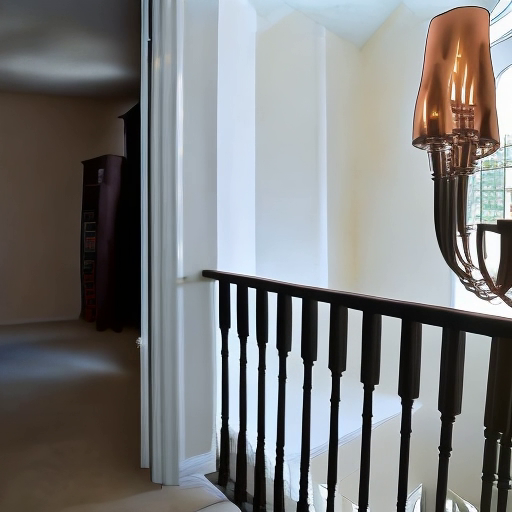}};
    \node[myNodeStyle] at ({7+\colgap},8.1) {\includegraphics[width=\qualImageSize]{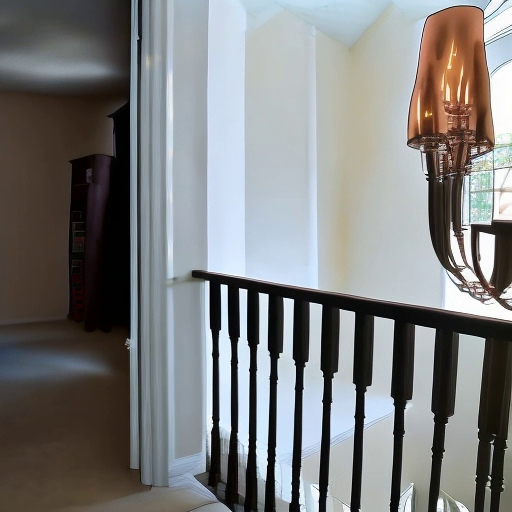}};
    \node[myNodeStyle] at ({8+\colgap},8.1) {\includegraphics[width=\qualImageSize]{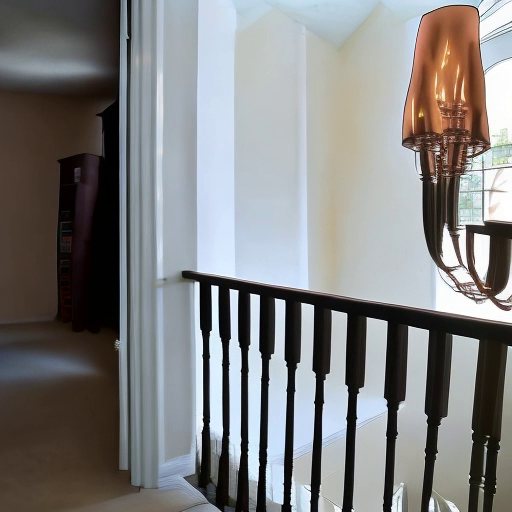}};

    \node[rotate=90] at (0.4,9.2) {\textbf{GT}};
    \node[myNodeStyle] at ({1+\colgap},9.2) {\includegraphics[width=\qualImageSize]{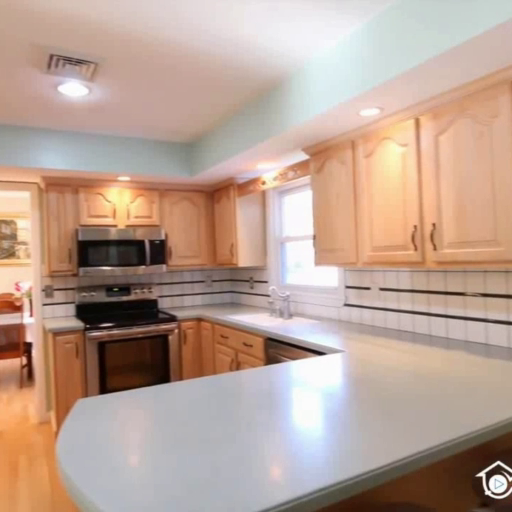}};
    \node[myNodeStyle] at ({2+\colgap},9.2) {\includegraphics[width=\qualImageSize]{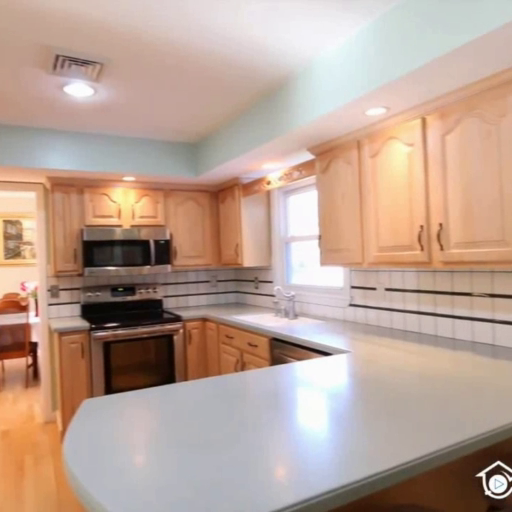}};
    \node[myNodeStyle] at ({3+\colgap},9.2) {\includegraphics[width=\qualImageSize]{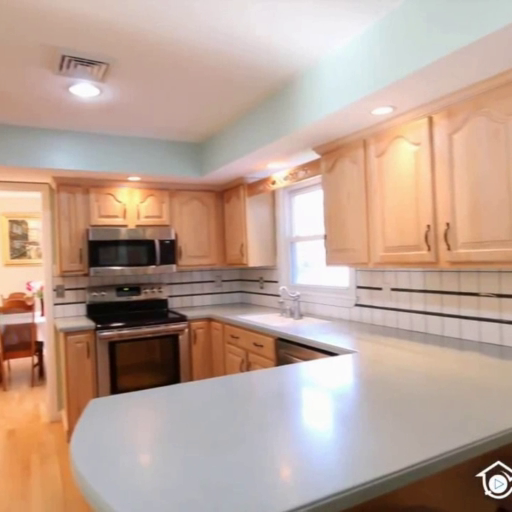}};
    \node[myNodeStyle] at ({4+\colgap},9.2) {\includegraphics[width=\qualImageSize]{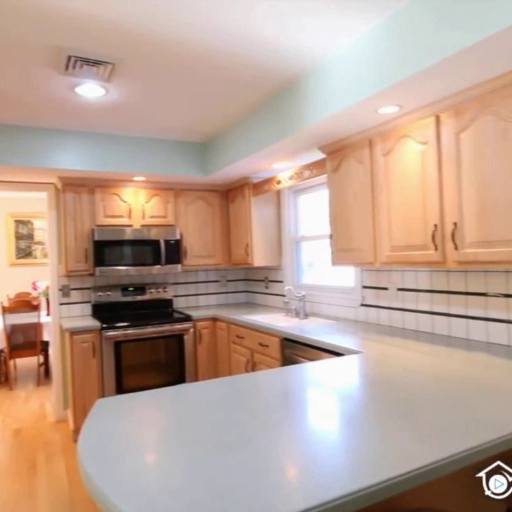}};
    \node[myNodeStyle] at ({5+\colgap},9.2) {\includegraphics[width=\qualImageSize]{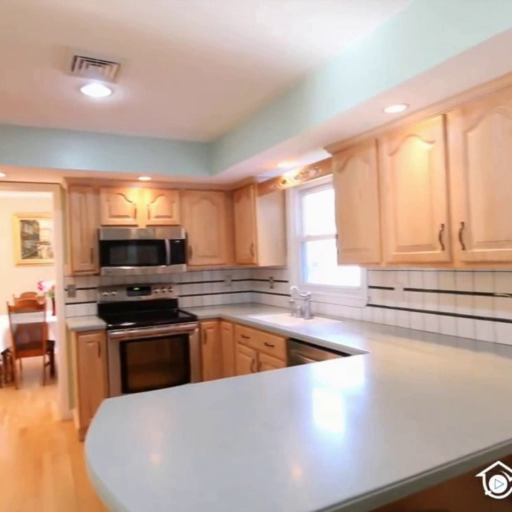}};
    \node[myNodeStyle] at ({6+\colgap},9.2) {\includegraphics[width=\qualImageSize]{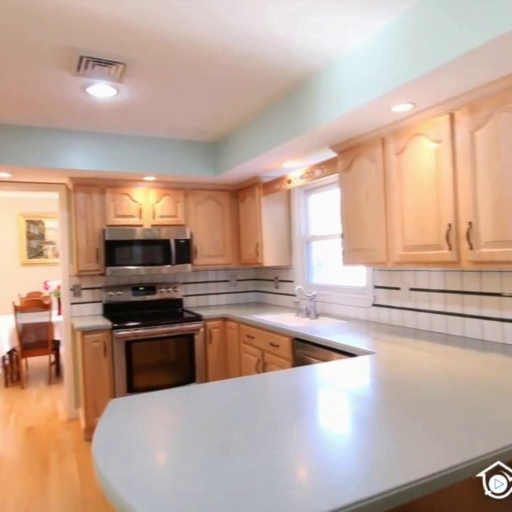}};
    \node[myNodeStyle] at ({7+\colgap},9.2) {\includegraphics[width=\qualImageSize]{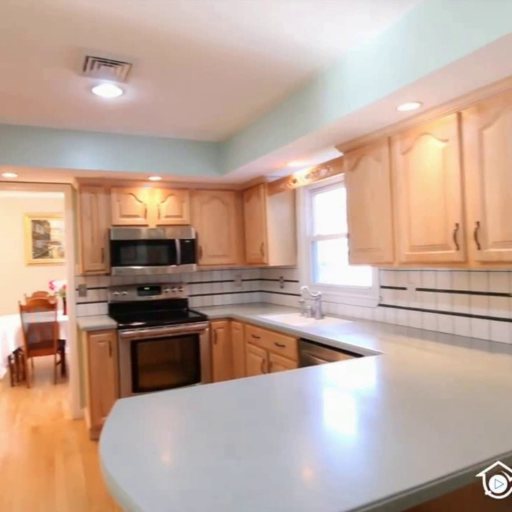}};
    \node[myNodeStyle] at ({8+\colgap},9.2) {\includegraphics[width=\qualImageSize]{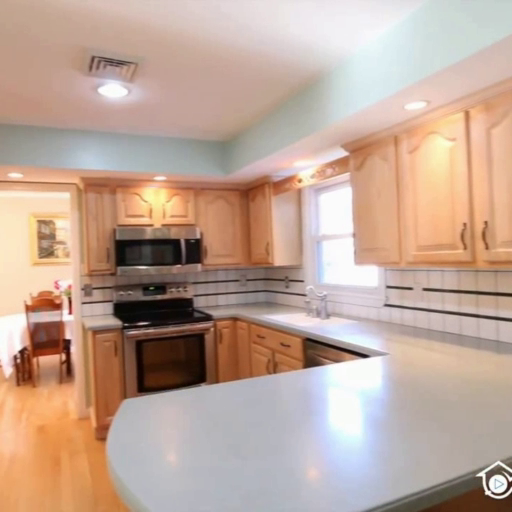}};

    \node[rotate=90] at (0.4,10.2) {\textbf{FrugalNeRF}};
    \node[myNodeStyle] at ({1+\colgap},10.2) {\includegraphics[width=\qualImageSize]{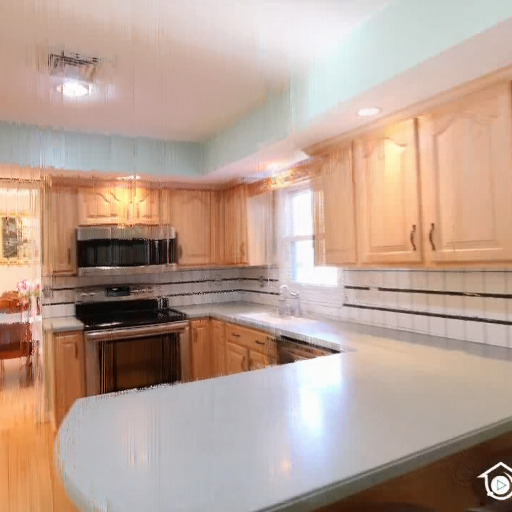}};
    \node[myNodeStyle] at ({2+\colgap},10.2) {\includegraphics[width=\qualImageSize]{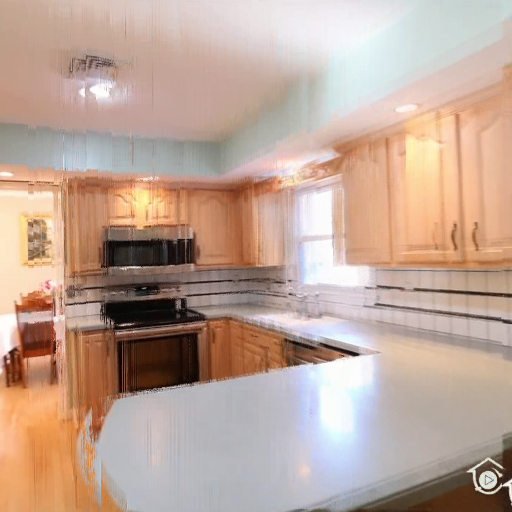}};
    \node[myNodeStyle] at ({3+\colgap},10.2) {\includegraphics[width=\qualImageSize]{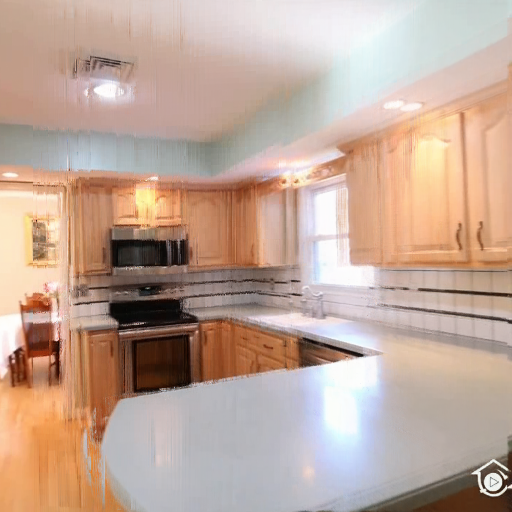}};
    \node[myNodeStyle] at ({4+\colgap},10.2) {\includegraphics[width=\qualImageSize]{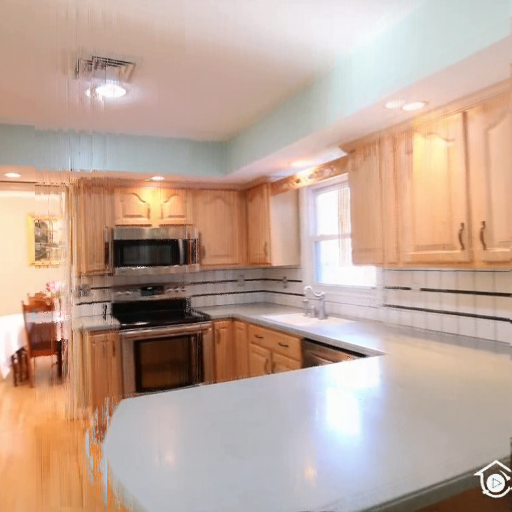}};
    \node[myNodeStyle] at ({5+\colgap},10.2) {\includegraphics[width=\qualImageSize]{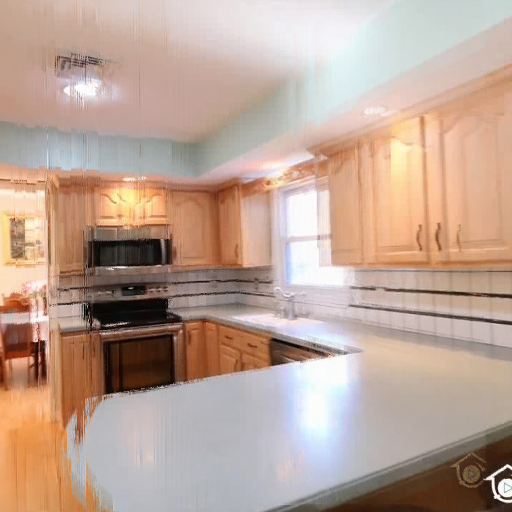}};
    \node[myNodeStyle] at ({6+\colgap},10.2) {\includegraphics[width=\qualImageSize]{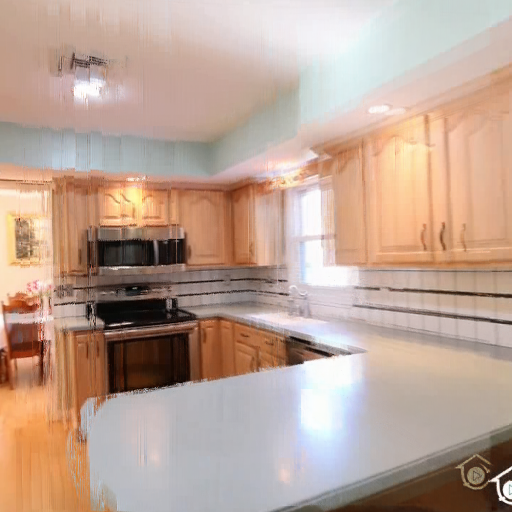}};
    \node[myNodeStyle] at ({7+\colgap},10.2) {\includegraphics[width=\qualImageSize]{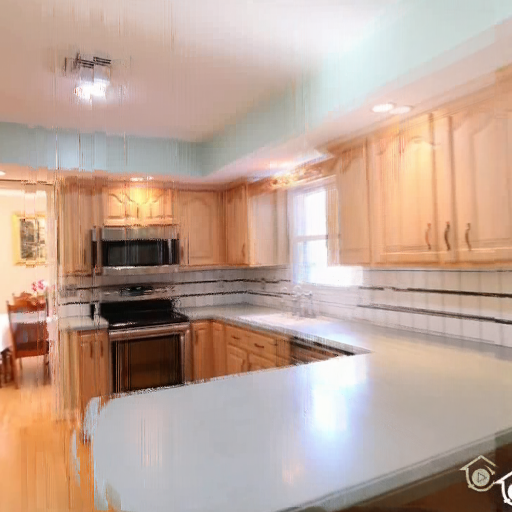}};
    \node[myNodeStyle] at ({8+\colgap},10.2) {\includegraphics[width=\qualImageSize]{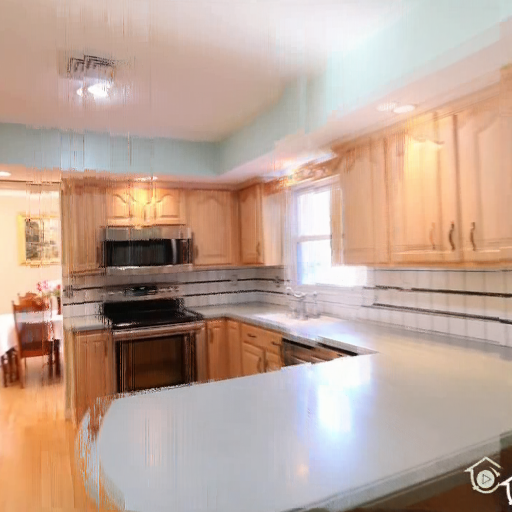}};

    \node[rotate=90] at (0.4,11.2) {\textbf{Ours}};
    \node[myNodeStyle] at ({1+\colgap},11.2) {\includegraphics[width=\qualImageSize]{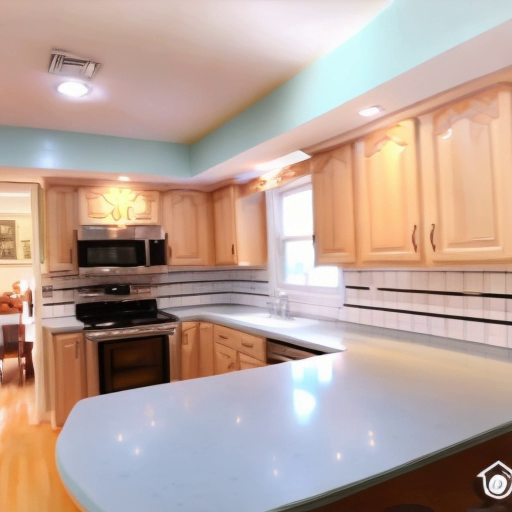}};
    \node[myNodeStyle] at ({2+\colgap},11.2) {\includegraphics[width=\qualImageSize]{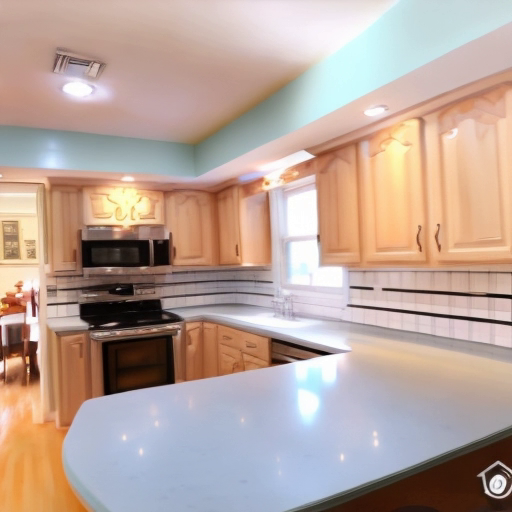}};
    \node[myNodeStyle] at ({3+\colgap},11.2) {\includegraphics[width=\qualImageSize]{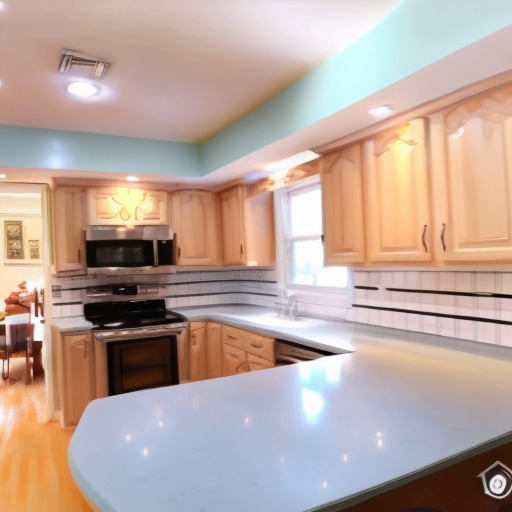}};
    \node[myNodeStyle] at ({4+\colgap},11.2) {\includegraphics[width=\qualImageSize]{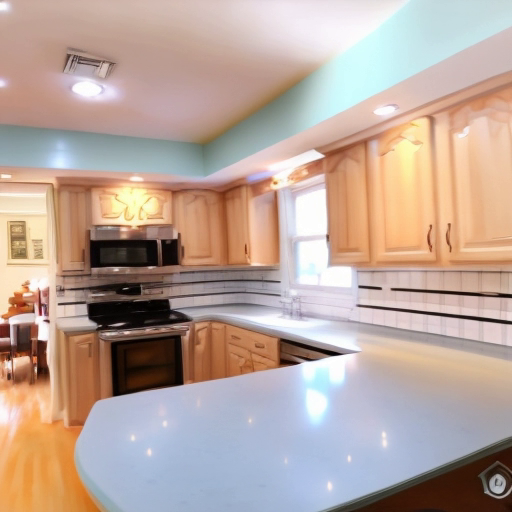}};
    \node[myNodeStyle] at ({5+\colgap},11.2) {\includegraphics[width=\qualImageSize]{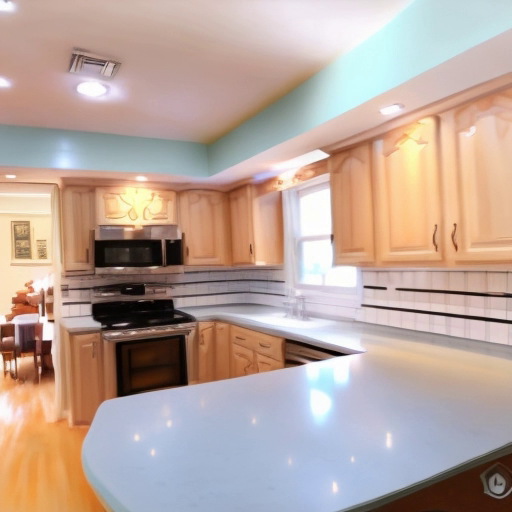}};
    \node[myNodeStyle] at ({6+\colgap},11.2) {\includegraphics[width=\qualImageSize]{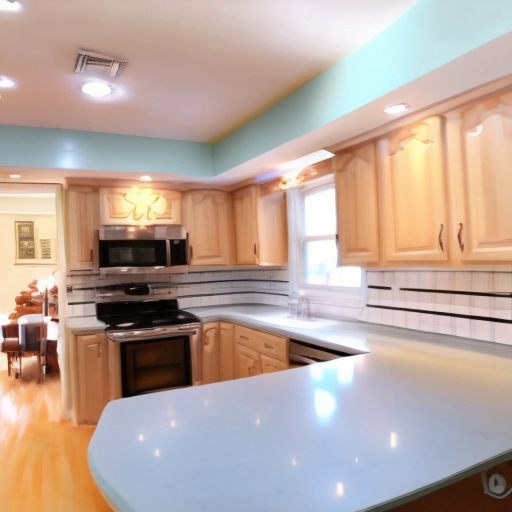}};
    \node[myNodeStyle] at ({7+\colgap},11.2) {\includegraphics[width=\qualImageSize]{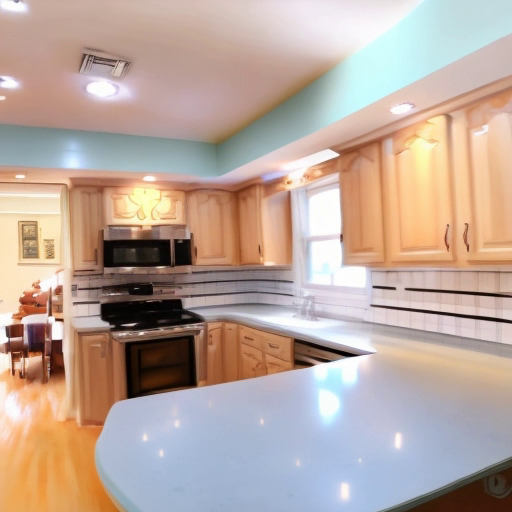}};
    \node[myNodeStyle] at ({8+\colgap},11.2) {\includegraphics[width=\qualImageSize]{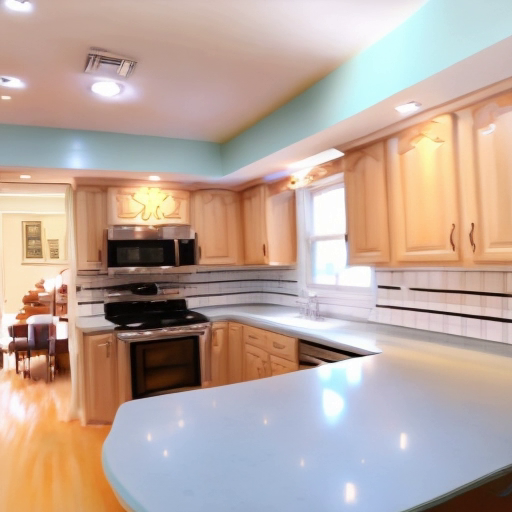}};

\end{tikzpicture}

%% file: main.bib
@String(IJCV   = {International Journal of Computer Vision})

@String(CVPR   = {IEEE Conference on Computer Vision and Pattern Recognition})

@String(ICCV   = {International Conference on Computer Vision})

@String(ECCV   = {European Conference on Computer Vision})

@String(NIPS   = {Advances in Neural Information Processing Systems})

@String(TOG    = {ACM Transactions on Graphics})

@String(ICLR   = {International Conference on Learning Representations})

@String(SIGGRAPH = {ACM Special Interest Group on Computer Graphics and Interactive Techniques Conference})

@String(MICCAI = {International Conference on Medical Image Computing and Computer Assisted Intervention})

@inproceedings{ddpm,
      title={Denoising Diffusion Probabilistic Models}, 
      author={Jonathan Ho and Ajay Jain and Pieter Abbeel},
      year={2020},
      journal=NIPS
}

@inproceedings{latent_diff,
      title={High-Resolution Image Synthesis with Latent Diffusion Models}, 
      author={Robin Rombach and Andreas Blattmann and Dominik Lorenz and Patrick Esser and Björn Ommer},
      year={2022},
      journal=CVPR
}

@inproceedings{videocrafter,
      title={{VideoCrafter1}: Open Diffusion Models for High-Quality Video Generation}, 
      author={Haoxin Chen and Menghan Xia and Yingqing He and Yong Zhang and Xiaodong Cun and Shaoshu Yang and Jinbo Xing and Yaofang Liu and Qifeng Chen and Xintao Wang and Chao Weng and Ying Shan},
      year={2023},
      journal={arXiv}
}

@inproceedings{stablevideodiffusion,
      title={{Stable Video Diffusion}: Scaling Latent Video Diffusion Models to Large Datasets}, 
      author={Andreas Blattmann and Tim Dockhorn and Sumith Kulal and Daniel Mendelevitch and Maciej Kilian and Dominik Lorenz and Yam Levi and Zion English and Vikram Voleti and Adam Letts and Varun Jampani and Robin Rombach},
      year={2023},
      journal={CoRR}
}

@inproceedings{seine,
      title={{SEINE}: Short-to-Long Video Diffusion Model for Generative Transition and Prediction}, 
      author={Xinyuan Chen and Yaohui Wang and Lingjun Zhang and Shaobin Zhuang and Xin Ma and Jiashuo Yu and Yali Wang and Dahua Lin and Yu Qiao and Ziwei Liu},
      year={2023},
      journal=ICLR
}

@inproceedings{controlnet,
      title={Adding Conditional Control to Text-to-Image Diffusion Models}, 
      author={Lvmin Zhang and Anyi Rao and Maneesh Agrawala},
      year={2023},
      journal=ICCV
}

@inproceedings{videodiffusion,
      title={Video Diffusion Models}, 
      author={Jonathan Ho and Tim Salimans and Alexey Gritsenko and William Chan and Mohammad Norouzi and David J. Fleet},
      year={2022},
      journal={arXiv},
}

@inproceedings{animatediff,
      title={{AnimateDiff}: Animate Your Personalized Text-to-Image Diffusion Models without Specific Tuning}, 
      author={Yuwei Guo and Ceyuan Yang and Anyi Rao and Zhengyang Liang and Yaohui Wang and Yu Qiao and Maneesh Agrawala and Dahua Lin and Bo Dai},
      year={2024},
      journal=ICLR
}

@inproceedings{direct_a_video, 
   title={{Direct-a-Video}: Customized Video Generation with User-Directed Camera Movement and Object Motion},
   author={Yang, Shiyuan and Hou, Liang and Huang, Haibin and Ma, Chongyang and Wan, Pengfei and Zhang, Di and Chen, Xiaodong and Liao, Jing},
   year={2024},
   journal=SIGGRAPH}

@inproceedings{cameractrl,
      title={{CameraCtrl}: Enabling Camera Control for Text-to-Video Generation}, 
      author={Hao He and Yinghao Xu and Yuwei Guo and Gordon Wetzstein and Bo Dai and Hongsheng Li and Ceyuan Yang},
      year={2024},
      journal=ICLR
}

@inproceedings{motionctrl,
      title={{MotionCtrl}: A Unified and Flexible Motion Controller for Video Generation}, 
      author={Zhouxia Wang and Ziyang Yuan and Xintao Wang and Tianshui Chen and Menghan Xia and Ping Luo and Ying Shan},
      year={2024},
      journal=SIGGRAPH
}

@inproceedings{cami2v,
      title={{CamI2V}: Camera-Controlled Image-to-Video Diffusion Model}, 
      author={Guangcong Zheng and Teng Li and Rui Jiang and Yehao Lu and Tao Wu and Xi Li},
      year={2024},
      eprint={2410.15957},
}

@inproceedings{xu2024camcocameracontrollable3dconsistentimagetovideo,
      title={{CamCo}: Camera-Controllable 3D-Consistent Image-to-Video Generation}, 
      author={Dejia Xu and Weili Nie and Chao Liu and Sifei Liu and Jan Kautz and Zhangyang Wang and Arash Vahdat},
      year={2024},
      journal={arXiv}
}

@inproceedings{dynamicrafter,
      title={{DynamiCrafter}: Animating Open-domain Images with Video Diffusion Priors}, 
      author={Jinbo Xing and Menghan Xia and Yong Zhang and Haoxin Chen and Wangbo Yu and Hanyuan Liu and Xintao Wang and Tien-Tsin Wong and Ying Shan},
      year={2023},
      journal=ECCV
}

@inproceedings{realestate10k,
      title={Stereo Magnification: Learning View Synthesis using Multiplane Images}, 
      author={Tinghui Zhou and Richard Tucker and John Flynn and Graham Fyffe and Noah Snavely},
      year={2018},
      journal=SIGGRAPH
}

@inproceedings{orbslam2,
   title={{ORB-SLAM2}: An Open-Source SLAM System for Monocular, Stereo, and RGB-D Cameras},
   journal={IEEE Transactions on Robotics},
   author={Mur-Artal, Raul and Tardos, Juan D.},
   year={2017} }

@inproceedings{fvd,
      title={Towards Accurate Generative Models of Video: A New Metric \& Challenges}, 
      author={Thomas Unterthiner and Sjoerd van Steenkiste and Karol Kurach and Raphael Marinier and Marcin Michalski and Sylvain Gelly},
      year={2019},
      eprint={arXiv},
}

@inproceedings{glomap,
      title={Global Structure-from-Motion Revisited}, 
      author={Linfei Pan and Dániel Baráth and Marc Pollefeys and Johannes L. Schönberger},
      year={2024},
      journal=ECCV
}

@inproceedings{lavie,
      title={{LAVIE}: High-Quality Video Generation with Cascaded Latent Diffusion Models}, 
      author={Yaohui Wang and Xinyuan Chen and Xin Ma and Shangchen Zhou and Ziqi Huang and Yi Wang and Ceyuan Yang and Yinan He and Jiashuo Yu and Peiqing Yang and Yuwei Guo and Tianxing Wu and Chenyang Si and Yuming Jiang and Cunjian Chen and Chen Change Loy and Bo Dai and Dahua Lin and Yu Qiao and Ziwei Liu},
      year={2023},
      journal=IJCV
}

@inproceedings{i2vgenxl,
      title={{I2VGen-XL}: High-Quality Image-to-Video Synthesis via Cascaded Diffusion Models}, 
      author={Shiwei Zhang and Jiayu Wang and Yingya Zhang and Kang Zhao and Hangjie Yuan and Zhiwu Qin and Xiang Wang and Deli Zhao and Jingren Zhou},
      year={2023},
      journal={arXiv}
}

@inproceedings{gen_l_video,
      title={{Gen-L-Video}: Multi-Text to Long Video Generation via Temporal Co-Denoising}, 
      author={Fu-Yun Wang and Wenshuo Chen and Guanglu Song and Han-Jia Ye and Yu Liu and Hongsheng Li},
      year={2023},
      journal={arXiv}
}

@inproceedings{videostudio,
      title={{VideoStudio}: Generating Consistent-Content and Multi-Scene Videos}, 
      author={Fuchen Long and Zhaofan Qiu and Ting Yao and Tao Mei},
      year={2024},
      journal=ECCV
}

@inproceedings{cogvideox,
      title={{CogVideoX}: Text-to-Video Diffusion Models with An Expert Transformer}, 
      author={Zhuoyi Yang and Jiayan Teng and Wendi Zheng and Ming Ding and Shiyu Huang and Jiazheng Xu and Yuanming Yang and Wenyi Hong and Xiaohan Zhang and Guanyu Feng and Da Yin and Xiaotao Gu and Yuxuan Zhang and Weihan Wang and Yean Cheng and Ting Liu and Bin Xu and Yuxiao Dong and Jie Tang},
      year={2025},
      journal=ICLR
}

@inproceedings{mevg,
      title={{MEVG}: Multi-event Video Generation with Text-to-Video Models}, 
      author={Gyeongrok Oh and Jaehwan Jeong and Sieun Kim and Wonmin Byeon and Jinkyu Kim and Sungwoong Kim and Sangpil Kim},
      year={2024},
      journal=ECCV
}

@inproceedings{vqgan,
      title={Taming Transformers for High-Resolution Image Synthesis}, 
      author={Patrick Esser and Robin Rombach and Björn Ommer},
      year={2021},
      journal=CVPR
}

@inproceedings{3dunet,
      title={{3D U-Net}: Learning Dense Volumetric Segmentation from Sparse Annotation}, 
      author={Özgün Çiçek and Ahmed Abdulkadir and Soeren S. Lienkamp and Thomas Brox and Olaf Ronneberger},
      year={2016},
      journal=MICCAI
}

@inproceedings{magicanimate,
      title={{MagicAnimate}: Temporally Consistent Human Image Animation using Diffusion Model}, 
      author={Zhongcong Xu and Jianfeng Zhang and Jun Hao Liew and Hanshu Yan and Jia-Wei Liu and Chenxu Zhang and Jiashi Feng and Mike Zheng Shou},
      year={2023},
      journal=CVPR
}

@inproceedings{disco,
      title={{DisCo}: Disentangled Control for Realistic Human Dance Generation}, 
      author={Tan Wang and Linjie Li and Kevin Lin and Yuanhao Zhai and Chung-Ching Lin and Zhengyuan Yang and Hanwang Zhang and Zicheng Liu and Lijuan Wang},
      year={2024},
      journal=CVPR
}

@inproceedings{stylecrafter,
      title={{StyleCrafter}: Enhancing Stylized Text-to-Video Generation with Style Adapter}, 
      author={Gongye Liu and Menghan Xia and Yong Zhang and Haoxin Chen and Jinbo Xing and Yibo Wang and Xintao Wang and Yujiu Yang and Ying Shan},
      year={2024},
      journal=TOG
}

@inproceedings{videocomposer,
      title={{VideoComposer}: Compositional Video Synthesis with Motion Controllability}, 
      author={Xiang Wang and Hangjie Yuan and Shiwei Zhang and Dayou Chen and Jiuniu Wang and Yingya Zhang and Yujun Shen and Deli Zhao and Jingren Zhou},
      year={2023},
      journal=NIPS
}

@inproceedings{stylemaster,
      title={{StyleMaster}: Stylize Your Video with Artistic Generation and Translation}, 
      author={Zixuan Ye and Huijuan Huang and Xintao Wang and Pengfei Wan and Di Zhang and Wenhan Luo},
      year={2024},
      journal={arXiv}
}

@inproceedings{ssim,
  author={Zhou Wang and Bovik, A.C. and Sheikh, H.R. and Simoncelli, E.P.},
  journal={IEEE Transactions on Image Processing}, 
  title={Image quality assessment: from error visibility to structural similarity}, 
  year={2004}}

@inproceedings{frugalnerf,
author = {Chin-Yang Lin and Chung-Ho Wu and Chang-Han Yeh and Shih-Han Yen and Cheng Sun and Yu-Lun Liu},
title = {{FrugalNeRF}: Fast Convergence for Extreme Few-shot Novel View Synthesis without Learned Priors},
journal = CVPR,
year = {2025}
}

@inproceedings{colmap,
    author={Sch\"{o}nberger, Johannes Lutz and Frahm, Jan-Michael},
    title={Structure-from-Motion Revisited},
    journal=CVPR,
    year={2016},
}

@inproceedings{bai2025recammaster,
  title={{ReCamMaster}: Camera-Controlled Generative Rendering from A Single Video},
  author={Bai, Jianhong and Xia, Menghan and Fu, Xiao and Wang, Xintao and Mu, Lianrui and Cao, Jinwen and Liu, Zuozhu and Hu, Haoji and Bai, Xiang and Wan, Pengfei and others},
  journal=ICCV,
  year={2025}
}

@inproceedings{Yu_2025_ICCV,
author = {Mark Yu and Wenbo Hu and Jinbo Xing and Ying Shan},
title = {{TrajectoryCrafter}: Redirecting Camera Trajectory for Monocular Videos via Diffusion Models},
journal = ICCV,
year = {2025}
}

@inproceedings{yan2021videogptvideogenerationusing,
author = {Wilson Yan and Yunzhi Zhang and Pieter Abbeel and Aravind Srinivas},
title = {{VideoGPT}: Video Generation using VQ-VAE and Transformers},
journal = ICLR,
year = {2021}
}

@inproceedings{karras2019stylebasedgeneratorarchitecturegenerative,
author = {Tero Karras and Samuli Laine and Timo Aila},
title = {A Style-Based Generator Architecture for Generative Adversarial Networks},
journal = CVPR,
year = {2019}
}
